\documentclass[runningheads]{llncs}
\usepackage{graphicx}
\usepackage{tikz}
\usepackage{comment}
\usepackage{amsmath,amssymb} % define this before the line numbering.
\usepackage{booktabs} % for professional tables
\usepackage{color}
\usepackage{hyperref}
\usepackage{cleveref}
\usepackage{adjustbox}
\usepackage{array,multirow,graphicx}
\usepackage{amsmath}
\usepackage{enumitem}
\usepackage{float}
\usepackage{calc}
\usepackage{siunitx}
\usepackage{makecell}

\usepackage[accsupp]{axessibility}  % Improves PDF readability for those with disabilities.

\definecolor{deepmagenta}{rgb}{0.8, 0.0, 0.8}

\definecolor{supervisedgray}{gray}{0.7}

\newcommand{\ourmethod}[0]{OWL-ViT}
\newcommand{\githuburl}{\href{https://github.com/google-research/scenic/tree/main/scenic/projects/owl\_vit}{\nolinkurl{github.com/google-research/scenic/tree/main/scenic/projects/owl\_vit}}}

\newcommand{\APr}[0]{$\text{AP}_\text{rare}$}
\newcommand{\cocoAP}[0]{$\text{AP}^\text{COCO}$}
\newcommand{\cocoAPfifty}[0]{$\text{AP50}^\text{COCO}$}
\newcommand{\objectsAP}[0]{$\text{AP}^\text{O365}$}
\newcommand{\objectsAPfifty}[0]{$\text{AP50}^\text{O365}$}
\newcommand{\oiAP}[0]{$\text{AP}^\text{OI}$}
\newcommand{\objectstreesixfiveAP}[0]{$\text{AP}^\text{O365}$}
\newcommand{\lvisAP}[0]{$\text{AP}^\text{LVIS}$}
\newcommand{\lvisAPr}[0]{$\text{AP}^\text{LVIS}_\text{rare}$}

\begin{document}
% \linenumbers
\pagestyle{headings}
\mainmatter

\title{Simple Open-Vocabulary Object Detection\\with Vision Transformers}

\titlerunning{Simple Open-Vocabulary Object Detection}
\author {
Matthias Minderer\thanks{Equal conceptual and technical contribution.} \and
Alexey Gritsenko\protect\footnotemark[1] \and \\
Austin Stone \and
Maxim Neumann \and
Dirk Weissenborn \and
Alexey Dosovitskiy \and
Aravindh Mahendran \and
Anurag Arnab \and
Mostafa Dehghani \and
Zhuoran Shen \and \\
Xiao Wang \and
Xiaohua Zhai \and
Thomas Kipf \and
Neil Houlsby
}
\authorrunning{Minderer et al.}
\institute{Google Research\\
\email{\{mjlm,agritsenko\}@google.com}}
%
%******************
\maketitle
\begin{abstract}
Combining simple architectures with large-scale pre-training has led to massive improvements in image classification.  
For object detection, pre-training and scaling approaches are less well established, especially in the long-tailed and open-vocabulary setting, where training data is relatively scarce. 
In this paper, we propose a strong recipe for transferring image-text models to open-vocabulary object detection.
We use a standard Vision Transformer architecture with minimal modifications, contrastive image-text pre-training, and end-to-end detection fine-tuning.
Our analysis of the scaling properties of this setup shows that increasing image-level pre-training and model size yield consistent improvements on the downstream detection task.
We provide the adaptation strategies and regularizations needed to attain very strong performance on zero-shot text-conditioned and one-shot image-conditioned object detection.
Code and models are available on GitHub\footnote{\githuburl{}}.

\keywords{open-vocabulary detection, transformer, vision transformer, zero-shot detection, image-conditioned detection, one-shot object detection, contrastive learning, image-text models, foundation models, CLIP}
\end{abstract}

\section{Introduction}
\label{sec:intro}

Object detection is a fundamental task in computer vision. Until recently, detection models were typically limited to a small, fixed set of semantic categories, because obtaining localized training data with large or open label spaces is costly and time-consuming. 
This has changed with the development of powerful language encoders and contrastive image-text training. These models learn a shared representation of image and text from loosely aligned image-text pairs, which are abundantly available on the web. By leveraging large amounts of image-text data, contrastive training has yielded major improvements in zero-shot classification performance and other language-based tasks \cite{radford2021clip,jia2021align,zhai2021lit}.

Many recent works aim to transfer the language capabilities of these models to object detection \cite{gu2021vild,li2021glip,zhong2021regionclip,zhou2022detic,kamath2021mdetr}. 
These methods, for example, use distillation against embeddings of image crops~\cite{gu2021vild}, weak supervision with image-level labels~\cite{zhou2022detic}, or self-training~\cite{li2021glip,zhong2021regionclip}.
Here, we provide a simple architecture and end-to-end training recipe that achieves strong open-vocabulary detection without these methods, even on categories not seen during training.

\begin{figure*}[t]
    \centerline{
        \includegraphics[width=1.0\textwidth]{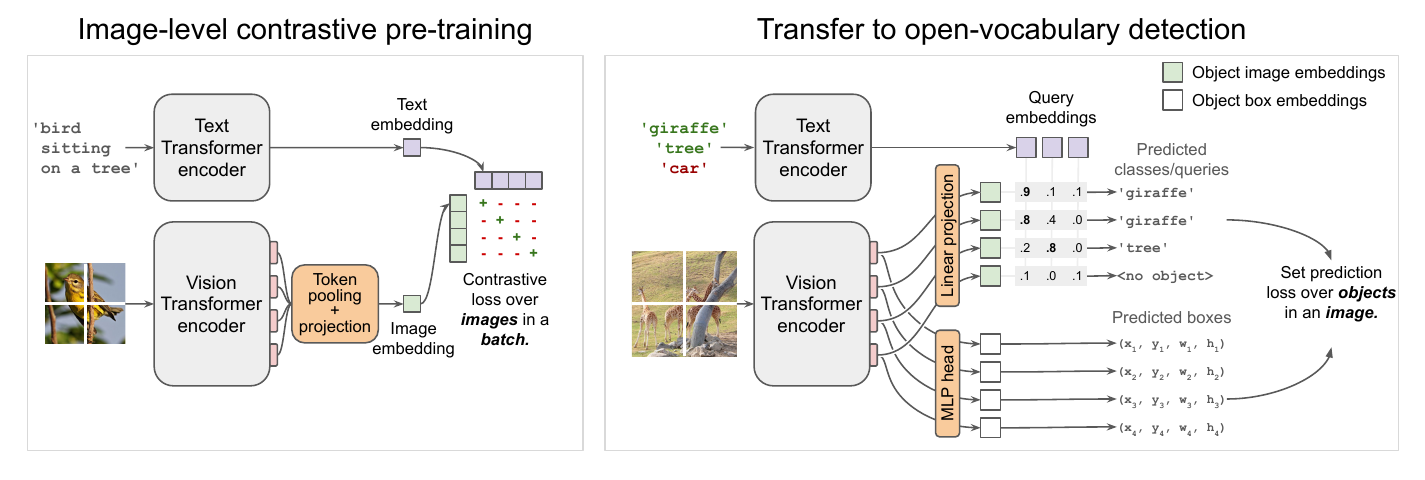}
    }
    \caption{
    Overview of our method. 
    \emph{Left:} We first pre-train an image and text encoder contrastively using image-text pairs, similar to CLIP~\cite{radford2021clip}, ALIGN~\cite{jia2021align}, and LiT~\cite{zhai2021lit}.
    \emph{Right:} We then transfer the pre-trained encoders to open-vocabulary object detection by removing token pooling and attaching light-weight object classification and localization heads directly to the image encoder output tokens. To achieve open-vocabulary detection, query strings are embedded with the text encoder and used for classification. The model is fine-tuned on standard detection datasets. At inference time, we can use text-derived embeddings for open-vocabulary detection, or image-derived embeddings for few-shot image-conditioned detection.
    }
    \label{fig:model_schematic}
\end{figure*}

We start with the Vision Transformer architecture~\cite{dosovitskiy2021vit}, which has been shown to be highly scalable, and pre-train it contrastively on a large image-text dataset~\cite{zhai2021lit,jia2021align}. To transfer the model to detection, we make a minimal set of changes: We remove the final token pooling layer and instead attach a lightweight classification and box head to each transformer output token. Open-vocabulary classification is enabled by replacing the fixed classification layer weights with the class-name embeddings obtained from the text model~\cite{bansal2018zeroshot} (\Cref{fig:model_schematic}). We fine-tune the pre-trained model on standard detection datasets using a bipartite matching loss \cite{carion2020detr}. Both the image and the text model are fine-tuned end-to-end.

We analyze the scaling properties of this approach and find that increasing model size and pre-training duration continue to yield improvements in detection performance beyond 20 billion image-text pairs. This is important since image-text pairs, in contrast to detection data, are abundant and allow further scaling.

A key feature of our model is its simplicity and modularity. Since the image and text components of our model are not fused, our model is agnostic to the source of query representations. We can therefore use our model without modification as a one-shot detection learner simply by querying it with image-derived embeddings. One-shot object detection is the challenging problem of detecting novel objects solely based on a query image patch showing the object~\cite{CoAE,laplacian,OS2D}. 
The image-conditioned one-shot ability is a powerful extension to text-conditioned detection because it allows detecting objects that are difficult to describe through text (yet easy to capture in an image), such as specialized technical parts. Despite using a generic architecture not specialized for this problem, we improve the state of the art for one-shot detection on \emph{unseen} 
COCO categories (held out during training) from 26.0 to 41.8 AP50, an improvement~of~72\%. 

For open-vocabulary text-conditioned detection, our model achieves 34.6\%~AP overall and 31.2\%~\APr{} on unseen classes on the LVIS dataset. % CLIP L/14 model.

In summary, we make the following contributions:
\begin{enumerate}
    \item A simple and strong recipe for transferring image-level pre-training to open-vocabulary object detection.
    \item State-of-the-art one-shot (image conditional) detection by a large margin.
    \item A detailed scaling and ablation study to justify our design.
\end{enumerate}

We believe our model will serve as a strong baseline that can be easily implemented in various frameworks, and as a flexible starting point for future research on tasks requiring open-vocabulary localization. We call our method \emph{Vision Transformer for Open-World Localization}, or \textit{\textbf{OWL-ViT}} for short.

\section{Related Work}
\label{sec:related}
\newcommand{\subsubshift}{\vspace{-4mm}}

\vspace{-2mm}
\subsubsection{Contrastive Vision-Language Pre-Training.} The idea of embedding images and text into a shared space has been used to achieve ``zero-shot'' generalization for a long time~\cite{frome2013devise,socher2013zero,xian2018zero}. Thanks to innovations in contrastive losses and better architectures, recent models can learn consistent visual and language representations from web-derived image and text pairs without the need for explicit human annotations. This vastly increases the available training data and has led to large improvements on zero-shot classification benchmarks~\cite{radford2021clip,jia2021align,zhai2021lit,basic}.
While any of the recent image-text models are compatible with our approach, our model and dataset are most similar to LiT \cite{zhai2021lit} and ALIGN \cite{jia2021align}. 

\subsubshift{}
\subsubsection{Closed-Vocabulary Object Detection.} Object detection models have been traditionally formulated for closed-vocabulary settings.
Initially, ``one-stage'' and ``two-stage'' detectors, such as SSD~\cite{liu2016ssd} and Faster-RCNN~\cite{ren2015faster} respectively, proliferated.
More recently, DETR~\cite{carion2020detr} showed that object detection can be framed as a set prediction problem, trained with bipartite matching, and achieve competitive results.
Notably, such architectures do not require region proposal generation or non-maximum suppression.
Follow-up works have proposed more efficient variants of DETR \cite{zhu2021deformable,yao2021efficient,song2022vidt}, including architectures without a ``decoder-stage''~\cite{fang2021you}.
Our work also simplifies DETR, in that we do not use a decoder.
Compared to~\cite{fang2021you}, which uses additional ``detection'' tokens, we further simplify the model by predicting one object instance directly from each image token.

\subsubshift{}
\subsubsection{Long-Tailed and Open-Vocabulary Object Detection.} To go beyond a closed vocabulary, fixed classification layers can be replaced by language embeddings to create open-vocabulary detectors~\cite{bansal2018zeroshot}. Open-vocabulary object detection has recently seen much progress from combining contrastively trained image-text models and classic object detectors~\cite{gu2021vild,kamath2021mdetr,li2021glip,zhong2021regionclip,zhou2022detic,zareian2021open}. 
The main challenge in this task is how to transfer the image-level representations of the image-text backbone to detection despite the scarcity of localized annotations for rare classes.
Making efficient use of the image-text pre-training is crucial since it allows for scaling without the need for expensive human annotations. 
Various approaches have been proposed. \textbf{ViLD}~\cite{gu2021vild} distills embeddings obtained by applying CLIP or ALIGN to cropped image regions from a class-agnostic region proposal network (RPN). The RPN, however, limits generalization performance on novel objects, which is exacerbated by ViLD's two-step distillation-training process. Multistage training is also used by \textbf{RegionCLIP}, which generates pseudo-labels on captioning data, followed by region-text contrastive pre-training, and transfer to detection. In contrast, our method fine-tunes both image and text models end-to-end on publicly available detection datasets, which simplifies training and improves generalization to unseen classes. \textbf{MDETR}~\cite{kamath2021mdetr} and \textbf{GLIP}~\cite{li2021glip} use a single text query for the whole image and formulate detection as the phrase grounding problem. This limits the number of object categories that can be processed per forward pass. Our architecture is simpler and more flexible in that it performs no image-text fusion and can handle multiple independent text or image-derived queries. \textbf{OVR-CNN}~\cite{zareian2021open} is most similar to our approach in that it fine-tunes an image-text model to detection on a limited vocabulary and relies on image-text pre-training for generalization to an open vocabulary. However, we differ in all modelling and loss function choices. We use ViT~\cite{dosovitskiy2021vit} instead of their ResNet~\cite{he2016resnet}, a DETR-like model instead of their Faster-RCNN~\cite{ren2015faster} and image-text pre-training as in LiT~\cite{zhai2021lit} instead of their PixelBERT~\cite{huang2020pixelbert} and visual grounding loss. Orthogonal to our approach, \textbf{Detic}~\cite{zhou2022detic} improves long-tail detection performance with weak supervision by training only the classification head on examples where only image-level annotations are available.

We note that in our definition of \emph{open-vocabulary} detection, object categories may overlap between detection training and testing. When we specifically refer to detecting categories for which no localized instances were seen during training, we use the term \emph{zero-shot}.

\subsubshift{}
\subsubsection{Image-Conditioned Detection.}
Related to open-vocabulary detection is the task of image-conditioned detection, which refers to the ability to detect objects matching a single \emph{query image} which shows an object of the category in question \cite{laplacian,CoAE,AIT,OS2D}. This task is also called \emph{one-shot object detection} because the query image is essentially a single training example. Image-based querying allows open-world detection when even the \emph{name} of the object is unknown, e.g. for unique objects or specialized technical parts. Our model can perform this task without modifications by simply using image-derived instead of text-derived embeddings as queries. Recent prior works on this problem have focused mainly on architectural innovations, for example using sophisticated forms of cross-attention between the query and target image \cite{CoAE,AIT}. Our approach instead relies on a simple but large model and extensive image-text pre-training.

\section{Method}

Our goal is to create a simple and scalable open-vocabulary object detector. 
We focus on standard Transformer-based models because of their scalability \cite{dosovitskiy2021vit} and success in closed-vocabulary detection \cite{carion2020detr}.
We present a two-stage recipe: 
\vspace{-0.5\topsep}
\begin{enumerate}[wide, labelwidth=0pt, labelindent=0pt]
    \item \makebox[0.9\linewidth][s]{Contrastively pre-train image and text encoders on large-scale image-text data.}
    \item Add detection heads and fine-tune on medium-sized detection data.
\end{enumerate}
\vspace{-0.5\topsep}
The model can then be queried in different ways to perform open-vocabulary or few-shot detection.

\subsection{Model}
\subsubsection{Architecture.}
\label{sec:method:architecture}

Our model uses a standard Vision Transformer as the image encoder and a similar Transformer architecture as the text encoder (\Cref{fig:model_schematic}).
To adapt the image encoder for detection, we remove the token pooling and final projection layer, and instead linearly project each output token representation to obtain per-object image embeddings for classification (\Cref{fig:model_schematic}, right).
The maximum number of predicted objects is therefore equal to the number of tokens (sequence length) of the image encoder. This is not a bottleneck in practice since the sequence length of our models is at least 576 (ViT-B/32 at input size $768 \times 768$), which is larger than the maximum number of instances in today's datasets (e.g., $294$~instances for LVIS~\cite{gupta2019lvis}).
Box coordinates are obtained by passing token representations through a small MLP.
Our setup resembles DETR~\cite{carion2020detr}, but is simplified by removing the decoder.

\subsubshift{}
\subsubsection{Open-vocabulary object detection.}

For open-vocabulary classification of detected objects, we follow prior work and use text embeddings, rather than learned class embeddings, in the output layer of the classification head \cite{bansal2018zeroshot}.
The text embeddings, which we call \emph{queries}, are obtained by passing category names or other textual object descriptions through the text encoder. 
The task of the model then becomes to predict, for each object, a bounding box and a probability with which each query applies to the object. 
Queries can be different for each image.
In effect, each image therefore has its own discriminative label space, which is defined by a set of text strings. This approach subsumes classical closed-vocabulary object detection as the special case in which the complete set of object category names is used as query set for each image.

In contrast to several other methods~\cite{li2021glip,kamath2021mdetr}, we do not combine all queries for an image into a single token sequence. Instead, each query consists of a separate token sequence which represents an individual object description, and is individually processed by the text encoder. In addition, our architecture includes no fusion between image and text encoders. 
Although early fusion seems intuitively beneficial, it dramatically reduces inference efficiency because encoding a query requires a forward pass through the entire image model and needs to be repeated for each image/query combination.
In our setup, we can compute query embeddings independently of the image, allowing us to use thousands of queries per image, many more than is possible with early fusion \cite{li2021glip}.

\subsubshift{}
\subsubsection{One- or Few-Shot Transfer.}
\label{sec:method:one-shot}
Our setup does not require query embeddings to be of textual origin. Since there is no fusion between image and text encoders, we can supply image- instead of text-derived embeddings as queries to the classification head without modifying the model. 
By using embeddings of prototypical object images as queries, our model can thus perform image-conditioned one-shot object detection. 
Using image embeddings as queries allows detection of objects which would be hard to describe in text.

\subsection{Training}

\subsubsection{Image-Level Contrastive Pre-Training.}

We pre-train the image and text encoder contrastively using the same image-text dataset and loss as in \cite{zhai2021lit} (\Cref{fig:model_schematic}, left).
We train both encoders from scratch with random initialization with a contrastive loss on the image and text representations.
For the image representation, we use multihead attention pooling (MAP)~\cite{lee2019map,vit-g} to aggregate token representation. 
The text representation is obtained from the final end-of-sequence (EOS) token of the text encoder. Alternatively, we use publicly available pre-trained CLIP models~\cite{radford2021clip} (details in \Cref{appendix:sec:hyperparamters}).

An advantage of our encoder-only architecture is that nearly all of the model's parameters (image and text encoder) can benefit from image-level pre-training. The detection-specific heads contain at most 1.1\% (depending on the model size) of the parameters of the model.

\subsubshift{}
\subsubsection{Training the Detector.}

Fine-tuning of pre-trained models for \emph{classification} is a well-studied problem. Classifiers, especially large Transformers, require carefully tuned regularization and data augmentation to perform well. Recipes for classifier training are now well established in the literature \cite{deit,steiner2021train,bello2021revisiting}. Here, we aim to provide a similar fine-tuning recipe for \emph{open-vocabulary detection}.

The general detection training procedure of our model is almost identical to that for closed-vocabulary detectors, except that we provide the set of object category names as queries for each image. The classification head therefore outputs logits over the per-image label space defined by the queries, rather than a fixed global label space.

We use the bipartite matching loss introduced by DETR \cite{carion2020detr}, but adapt it to long-tailed/open-vocabulary detection as follows. Due to the effort required for annotating detection datasets exhaustively, datasets with large numbers of classes are annotated in a federated manner \cite{gupta2019lvis,kuznetsova2020openimages}. Such datasets have non-disjoint label spaces, which means that each object can have multiple labels. We therefore use focal sigmoid cross-entropy \cite{zhu2021deformable} instead of softmax cross-entropy as the classification loss. Further, since not all object categories are annotated in every image, federated datasets provide both positive (present) and negative (known to be absent) annotations for each image. During training, for a given image, we use all its positive and negative annotations as queries. Additionally, we randomly sample categories in proportion to their frequency in the data and add them as ``pseudo-negatives'' to have at least 50 negatives per image~\cite{zhou2021probabilistic}.

Even the largest federated detection datasets contain only $\approx10^6$ images, which is small in contrast to the billions of image-level weak labels which exist for pre-training \cite{instagram,vit-g,radford2021clip,jia2021align}. It is known that large Transformers trained on datasets of this size (such as ImageNet-1k) require carefully-tuned regularization and data augmentation to perform well \cite{deit,steiner2021train,bello2021revisiting}. We found the same to be true for detection training and provide a detailed breakdown of the augmentations and regularizations required to achieve very high performance with large Transformers in \Cref{sec:experiments:unlock-downstream-potential}. 

\section{Experiments}
\label{sec:experiments}

\subsection{Model Details}

For the image model, we use standard Vision Transformers \cite{dosovitskiy2021vit}. We follow the nomenclature from~\cite{dosovitskiy2021vit} for model size, patch size, and Transformer vs. hybrid architectures. For example, B/32 refers to ViT-Base with patch size 32, while R50+H/32 refers to a hybrid ResNet50 + ViT-Huge with stride 32. 

For the text model, we use a Transformer architecture similar to the image model. Unless otherwise noted, we use a text model with 12 layers, 512 hidden size ($D$), 2048 MLP size and 8 heads (this is smaller than B). 

Image and text models are first pre-trained on the image level and then fine-tuned on object-level annotations. Pre-training is performed from scratch as in LiT~\cite{zhai2021lit} (\texttt{uu} in their notation) on their dataset of 3.6 billion image-text pairs. 

After pre-training, token pooling is removed and detection heads are added (see \Cref{sec:method:architecture} and \Cref{fig:model_schematic}). The model predicts one box for each output token. We add a bias to the predicted box coordinates such that each box is by default centered on the image patch that corresponds to the token from which this box is predicted when arranging the token sequence as a 2D grid. The model therefore predicts the difference from that default location, similar to how Region Proposal Networks~\cite{ren2015faster} predict offsets with respect to pre-defined anchors. Although there is no strict correspondence between image patches and tokens representations later in the Transformer network, biasing box predictions in this way speeds up training and improves final performance (\Cref{sec:experiments:unlock-downstream-potential}).

We use an image size of $224 \times 224$ in most models for pre-training (see Appendix~\ref{appendix:sec:hyperparamters}) and larger sizes for detection fine-tuning and evaluation (specified in \Cref{tab:main-results}). To change model input size after pre-training, we resize the image position embeddings with linear interpolation.
Models are fine-tuned at a batch size of 256 for at most 140'000 steps (fewer for larger models).
We implement our model using JAX~\cite{jax2018github} and the \emph{Scenic} library~\cite{scenic}.

\subsection{Detection Data}
\label{sec:detection-data}
Due to the open-vocabulary design of our model, we can easily combine datasets with different label spaces by replacing integer labels with class name strings. For object-level training, we use publicly available detection datasets with a total of around 2~million images (OpenImages~V4 (OI)~\cite{kuznetsova2020openimages}, Objects~365 (O365) \cite{shao2019vg}, and/or Visual Genome (VG)~\cite{krishnavisualgenome}, as indicated). Evaluation is performed on the COCO \cite{lin2014coco}, LVIS \cite{gupta2019lvis}, and O365. For dataset details, see Appendix~\ref{appendix:sec:datasets}.

Since OI, VG, O365 and the image-level pre-training data contain images that are also in COCO / LVIS, we use a strict deduplication procedure to remove any COCO or LVIS test and validation images from all datasets we use for training (see  Appendix~\ref{appendix:sec:datasets-dedup} for details). Unless otherwise noted, we mix OI and VG randomly at a ratio of $70\%$ to $30\%$ for detection training in our experiments. In~\Cref{tab:main-results}, as indicated, we use either LVIS base training (for comparability to prior work), or O365 and VG at a ratio of $80\%$ to $20\%$. We use a range of image and label augmentations, which we discuss in \Cref{sec:experiments:unlock-downstream-potential}.

\subsection{Open-Vocabulary Detection Performance}

\begingroup
\setlength{\tabcolsep}{1mm}
\begin{table}[t]
    \newcommand{\superv}[1]{\textcolor{supervisedgray}{#1}}
    \renewcommand{\bf}[1]{\textbf{#1}}
    \centering
    \caption{Open-vocabulary and zero-shot performance on LVIS v1.0 val. 
    For our models, we remove annotations matching LVIS rare category names from all detection training datasets, such that \lvisAPr{} measures zero-shot performance.
    \superv{Gray} numbers indicate models trained on the LVIS frequent and common (``base'') annotations.
    For reference, ViT-B/32 is comparable to ResNet50 in inference compute (139.6 vs 141.5 GFLOPs).
    For our models, we report the mean performance over three fine-tuning runs.
    Results for COCO and O365 are provided in \Cref{appendix:sec:coco-o365-results}.
    }
    \label{tab:main-results}
    \resizebox{\textwidth}{!}{%
    \begin{tabular}{cllccccc}
    \toprule
     & \bf{Method}                             & \bf{Backbone}         & \bf{Image-level}      & \bf{Object-level}         & \bf{Res.}         & \bf{\lvisAP{}}    & \bf{\lvisAPr{}}    \\
    \midrule
    \multicolumn{8}{l}{\textbf{\textit{LVIS base training:}}}\\
    
    % ViLD: https://arxiv.org/pdf/2104.13921.pdf Table 3
    \hspace{-1mm}1 & ViLD-ens \cite{gu2021vild}              & ResNet50              & CLIP                  & LVIS base                 & 1024              & \superv{25.5}   & 16.6        \\
    \hspace{-1mm}2 & ViLD-ens \cite{gu2021vild}              & EffNet-b7             & ALIGN                 & LVIS base                 & 1024              & \superv{29.3}   & 26.3        \\
    
    % Reg.CLIP: https://arxiv.org/pdf/2112.09106.pdf Table 2
    \hspace{-1mm}3 & Reg.~CLIP \cite{zhong2021regionclip}     & R50-C4                & CC3M                  & LVIS base                 & ?                 & \superv{28.2}   & 17.1        \\
    \hspace{-1mm}4 & Reg.~CLIP \cite{zhong2021regionclip}     & R50x4-C4              & CC3M                  & LVIS base                 & ?                 & \superv{32.3}   & 22.0        \vspace{1.5mm}\\
    \hspace{-1mm}5 & \ourmethod{} (ours)                     & ViT-H/14              & LiT                   & LVIS base                & 840               & \superv{35.3}            & 23.3        \\
    
    \hspace{-1mm}6 & \ourmethod{} (ours)                     & ViT-L/14      & CLIP                  & LVIS base                    & 840               & \superv{34.7}              & 25.6         \\
    
    \midrule
    \multicolumn{8}{l}{\textbf{\textit{Unrestricted open-vocabulary training:}}}\\
    
    % GLIP: https://arxiv.org/pdf/2112.03857.pdf Table 4
    \hspace{-1mm}7 & GLIP \cite{li2021glip}                  & Swin-T                & Cap4M                 & O365, GoldG, ...          & ?                 & 17.2            & 10.1        \\
    \hspace{-1mm}8 & GLIP \cite{li2021glip}                  & Swin-L                & CC12M, SBU            & OI, O365,  VG, ...        & ?                 & 26.9            & 17.1        \vspace{1.5mm}\\
    
    % Ours:
    \hspace{-1mm}9 & \ourmethod{} (ours)                     & ViT-B/32              & LiT                   & O365, VG                    & 768               & 23.3            & 19.7        \\
    \hspace{-1mm}11 & \ourmethod{} (ours)                     & R26+B/32              & LiT                   & O365, VG                    & 768               & 25.7            & 21.6        \\
    \hspace{-1mm}10 & \ourmethod{} (ours)                     & ViT-B/16              & LiT                   & O365, VG                    & 768               & 26.7            & 23.6        \\
    \hspace{-1mm}12 & \ourmethod{} (ours)                     & ViT-L/16              & LiT                   & O365, VG                    & 768               & 30.9            & 28.8        \\
    \hspace{-1mm}13 & \ourmethod{} (ours)                     & ViT-H/14              & LiT                   & O365, VG              & 840               & 33.6            & 30.6       \vspace{1.5mm}\\
    
    % Ours + CLIP:
    \hspace{-1mm}14 & \ourmethod{} (ours)                     & ViT-B/32      & CLIP                  & O365, VG                    & 768               & 22.1              & 18.9        \\
    \hspace{-1mm}15 & \ourmethod{} (ours)                     & ViT-B/16      & CLIP                  & O365, VG                    & 768               & 27.2              & 20.6          \\
    \hspace{-1mm}16 & \ourmethod{} (ours)                     & ViT-L/14      & CLIP                  & O365, VG                    & 840               & 34.6              & 31.2          \\
    \bottomrule
    \end{tabular}}
\end{table}
\endgroup

We use LVIS~v1.0~val~\cite{gupta2019lvis} as our main benchmark since this dataset has a long tail of rare categories and is therefore well-suited to measure open-vocabulary performance.
For evaluation, we use all category names as query for each image, i.e. 1203 queries per image for LVIS.
Class predictions are ensembled over seven prompt templates as described in \Cref{sec:experiments:unlock-downstream-potential}.
Some LVIS categories appear in the datasets we use for training. To measure performance on unseen categories, we therefore remove from our training data all box annotations with labels that match any of the LVIS ``rare'' categories. The \lvisAPr{} metric therefore measures the ``zero-shot'' performance of our model in the sense that the model has not seen localized annotations for these categories.

\Cref{tab:main-results} shows LVIS results for our models and a range of prior work. We compare to open-vocabulary models that do not train on the full LVIS dataset. Results obtained by training on parts of LVIS (e.g. ``base'' categories~\cite{gu2021vild}) are shown in gray. Our method is highly competitive across architecture sizes in both open-vocabulary (\lvisAP{}) and zero-shot (\lvisAPr{}) scenarios. Our best model achieves 31.2\% \lvisAPr{} and uses a publicly available CLIP backbone.

For comparison to prior work, we also provide results on MS-COCO~2017 and Objects~365. For these evaluations, we train models on OI+VG instead of O365+VG, to measure generalization. However, most COCO and O365 categories are present in the training data and we do not remove them, since they constitute a large fraction of the available annotations. Our COCO and O365 results are therefore not ``zero-shot'', but test the open-vocabulary transfer ability of our model. Our best model (CLIP L/14; see \Cref{tab:main-results}) achieves 43.5\%~\cocoAP{}; a version of the model trained without O365 achieves 15.8\%~\objectstreesixfiveAP{} (further results in \Cref{appendix:sec:coco-o365-results}).

\begin{figure}[t]
    \centering
    \includegraphics[width=1.0\textwidth]{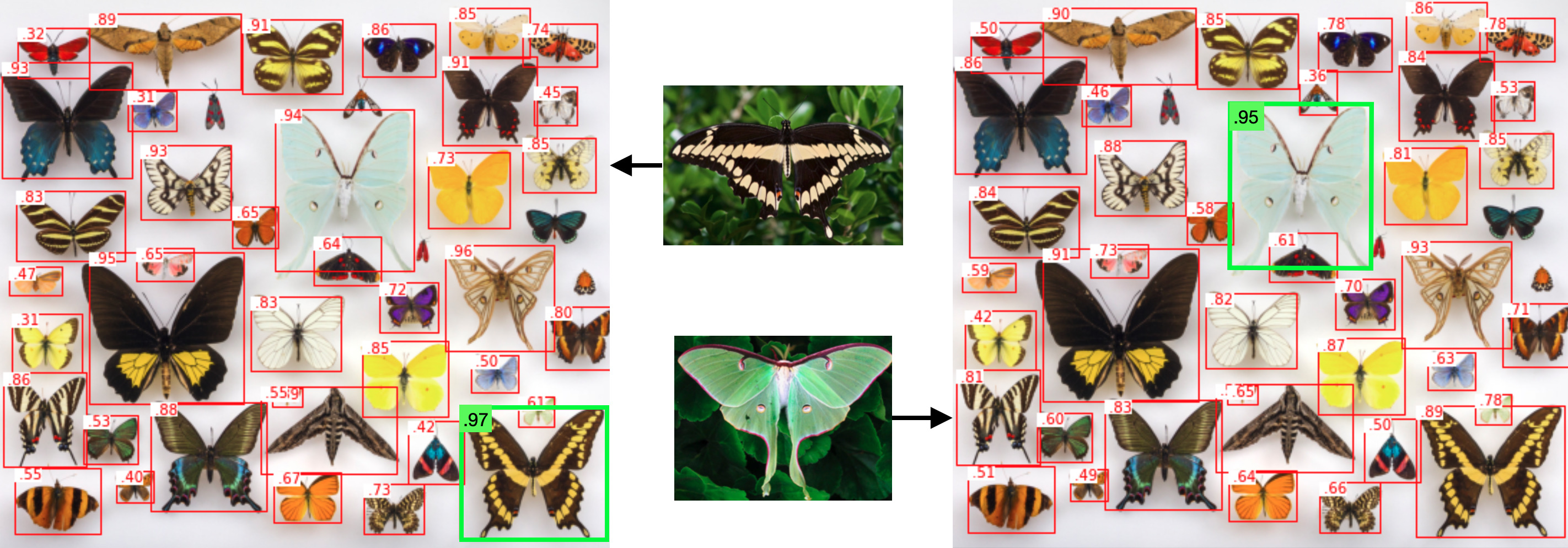}
    \vspace{-5mm}
    \caption{
    Example of one-shot image-conditioned detection. Images in the middle are used as queries; the respective detections on the target image are shown on the left and right. In both cases, the highest score is given to instances of the species matching the query. In contrast, text-based querying (not shown) detects the correct species only for the top example (``swallowtail butterfly'') but not for the bottom (``luna moth'').
    }
    \label{fig:image-conditioning-butterflies}
\end{figure}

\subsection{Few-Shot Image-Conditioned Detection Performance}
\label{sec:one-shot}

As described in \Cref{sec:method:one-shot}, our model can perform one- or few-shot object detection simply be replacing text-derived query embeddings with image-derived query embeddings. In few-shot detection, we are given a query image with a box around an example object. The goal is to detect objects of the same category as the example in new target images. To get the query embedding, we first run inference on the query image and select a predicted detection which has high box overlap with the query box (after some filtering; see Appendix~\ref{appendix:sec:image-conditioning-details} for details). We then use the image embedding of that prediction as query on the test images. 

For evaluation on this task, we follow the procedure described in \cite{CoAE}: During detection training, we hold out some COCO categories to evaluate on, and in addition all synonymous and semantically descendant categories that appear in our detection training data. We do not modify the image-text pre-training stage. 

Despite not being designed specifically for this task, our model strongly outperforms the best task-specific prior work by a margin of $72$\% across the four COCO splits as shown in~\Cref{tab:coco_one_shot}. Unlike prior work, our model does not entangle query image and target image features during inference, which enables us to run our models on thousands of different image embeddings simultaneously and efficiently, enhancing its practicality.

To move beyond a single query example (one-shot) to few-shot predictions, we can simply average image embeddings for multiple query examples for each category. This leads to further significant improvements (\Cref{tab:coco_one_shot}, bottom row).

\newcommand{\STAB}[1]{\begin{tabular}{@{}c@{}}#1\end{tabular}}
\begin{table}[t]
 \setlength{\tabcolsep}{8pt}
  \centering
  \caption{One- and few-shot image-conditioned detection performance on COCO~AP50. Our method (R50+H/32 architecture) strongly outperforms prior work and also shows marked improvements as the number of conditioning queries is increased to $k=10$. COCO category splits as in \cite{CoAE}. Because the evaluation is stochastic, for our results, we report the average across $3$~runs.}
  \label{tab:coco_one_shot}
  \resizebox{0.9\textwidth}{!}{%
  \begin{tabular}{clccccc}
    \toprule
     & \multicolumn{1}{c}{\textbf{Method}} & \multicolumn{1}{c}{\textbf{Split  1}} & \multicolumn{1}{c}{\textbf{Split  2}} & \multicolumn{1}{c}{\textbf{Split  3}} & \multicolumn{1}{c}{\textbf{Split  4}}  & \multicolumn{1}{c}{\textbf{Mean}} \\
     \midrule
     \multirow{5}{*}{\STAB{\rotatebox[origin=c]{90}{Seen}}}
     & SiamMask~\cite{siamMask}             &38.9                        &37.1                       &37.8                    &36.6                            &37.6 \\
     & CoAE~\cite{CoAE}                     &42.2                        &40.2                       &39.9                    &41.3                            &40.9                           \\
     & AIT~\cite{AIT}                      &\textbf{50.1}                        &47.2                        &45.8                    &46.9                            &47.5                           \\
     & OWL-ViT~(ours) & 49.9 & \textbf{49.1}  &\textbf{49.2}   & \textbf{48.2} &\textbf{49.1} \\
     \cmidrule{2-7}
     & OWL-ViT~($k=10$; ours)       & 54.1                       & 55.3                            & 56.2                       & 54.9                               & 55.1      \\
     \midrule

     \multirow{5}{*}{\STAB{\rotatebox[origin=c]{90}{Unseen}}}
     & SiamMask~\cite{siamMask}             &15.3                        &17.6                       &17.4                       &17.0                          &16.8 \\
     & CoAE~\cite{CoAE}                     &23.4                        &23.6                       &20.5                       &20.4                          &22.0                           \\
     & AIT~\cite{AIT}                      &26.0                        &26.4                       &22.3                       &22.6                          &24.3 \\
     & OWL-ViT~(ours) &\textbf{43.6}  &\textbf{41.3} &\textbf{40.2} &\textbf{41.9} &\textbf{41.8}\\ \cmidrule{2-7}
     & OWL-ViT~($k=10$; ours)      &49.3              &51.1               & 42.4               & 44.5                 & 46.8 \\
     \bottomrule
  \end{tabular}}
\end{table} 

\subsection{Scaling of Image-Level Pre-Training}
\label{sec:scaling}

\begin{figure*}[ht!]
    \centerline{
        \includegraphics[width=0.3\textwidth]{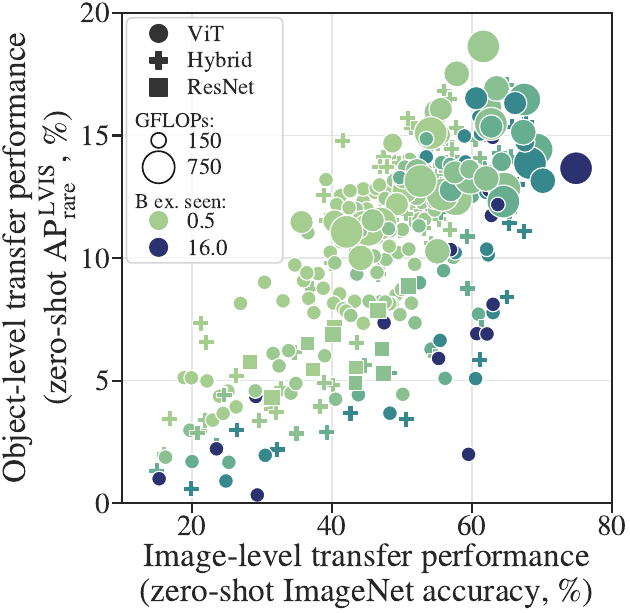}
        \hfill
        \includegraphics[width=0.64\textwidth,trim=0 -10mm 0 0]{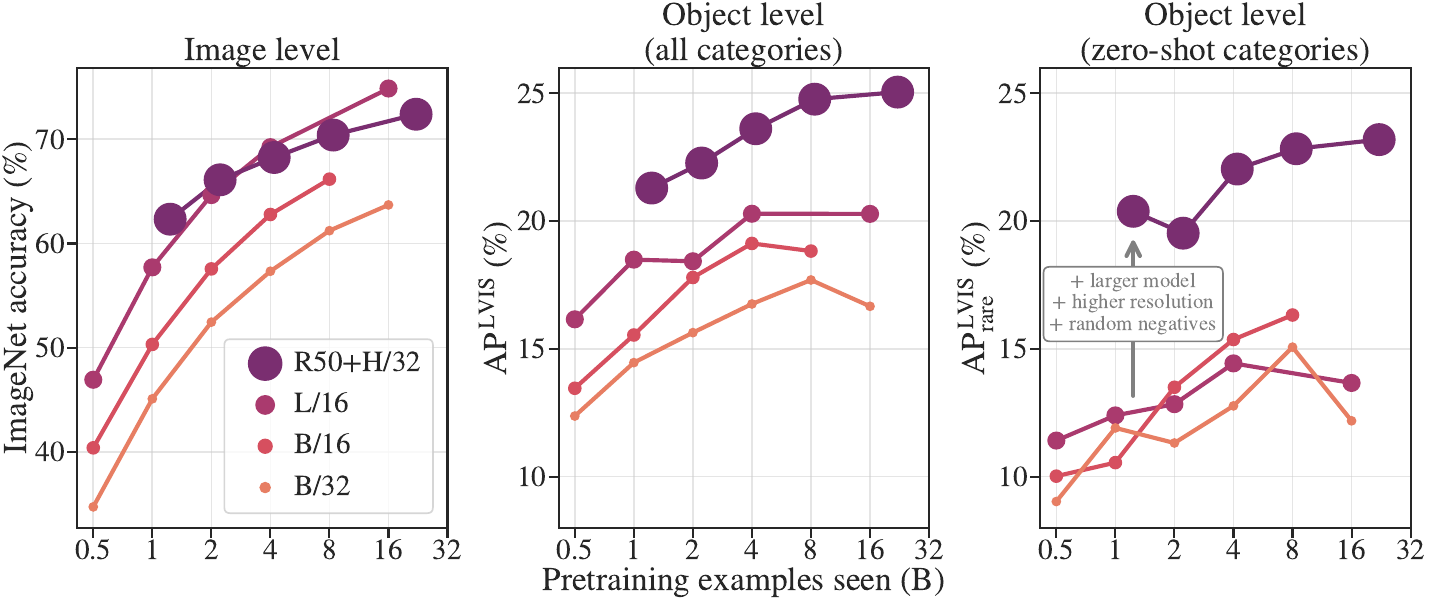}
    }
    \caption{
    Image-level pre-training transfers to detection.
    \emph{Left:} Overview of the relationship between image-level performance (zero-shot ImageNet accuracy after pre-training) and object-level performance (\lvisAPr{} after detection fine-tuning) of contrastively trained image-text models. Each dot represents one pre-training configuration and its best detection performance across a range of learning rates and weight decays. Configurations vary in encoder architecture (ViT/Hybrid/ResNet), model size (in order of detection inference compute: R50, B/32, R26+B/32, R101, L/32, B/16, H/32, R50+H/32, L/16), and pre-training duration (billions of examples seen including repetitions; 3.6B unique examples). 
    High image-level performance is necessary, but not sufficient, for high object-level performance (Pearson's $r = 0.73$; in contrast, image-level transfer performance correlates better with pre-training-task performance: $r = 0.98$). 
    \emph{Right:} Across model sizes, longer image-level pre-training translates to higher object-level performance. Further gains on detection are possible by scaling up fine-tuning.
    }
    \label{fig:image_vs_object_level_performance}
\end{figure*}

After establishing that our method achieves strong open-vocabulary, zero-shot, and image-conditioned detection performance, we next analyze its scaling properties and design choices. We focus on image-level pre-training in this section. In \Cref{sec:experiments:unlock-downstream-potential}, we will describe the fine-tuning methods that are necessary for successful transfer of the pre-trained model to detection.

To understand how image-level pre-training relates to final detection performance, we systematically explored the dimensions of pre-training duration, model size, and model architecture. For every configuration, we pre-trained and then fine-tuned several models across a range of learning rates and weight decays, since the optimal settings of these parameters vary by configuration (see Appendix~\ref{appendix:sec:hyperparamters} for a list of covered settings).

We first consider how well image-level pre-training transfers to detection in general. \Cref{fig:image_vs_object_level_performance} shows the relationship between image-level performance (zero-shot ImageNet accuracy) and object-level performance (zero-shot \lvisAPr{}) for all architecture, size, and pre-training-duration configurations covered by our study (the best result across learning rates and weight decays is shown). We find that, while the best object-level models typically also have good image-level performance, the reverse is not true: many models that do well to the image-level task transfer poorly to detection. In other words, high image-level performance is necessary, but not sufficient, for strong transfer to detection.

\begin{figure*}[thb!]
    \includegraphics[width=0.64\textwidth]{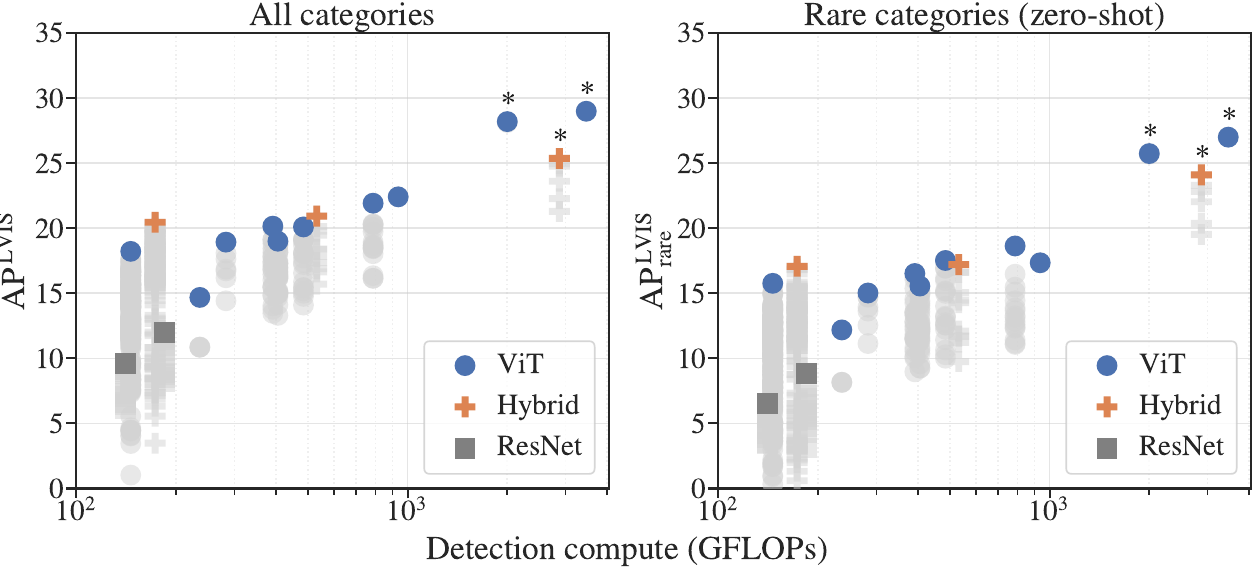}
    \hfill
    \includegraphics[width=0.30\textwidth,trim=0 -6mm 0 0]{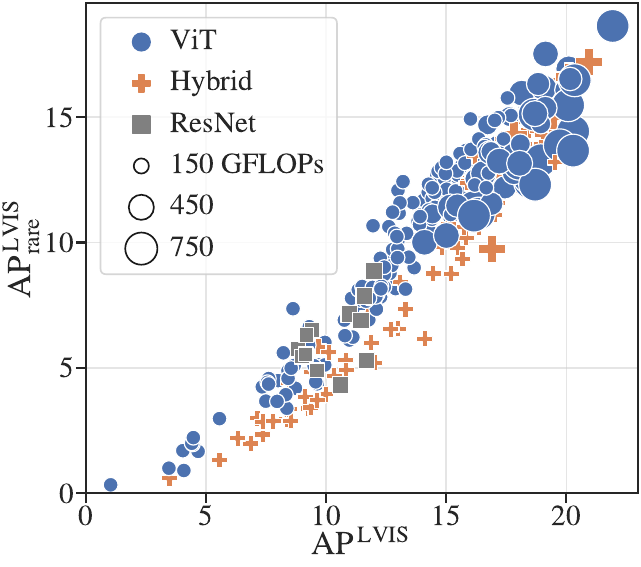}
    \caption{
    Effect of model architecture on detection performance.
    \emph{Left:} Hybrid architectures are more efficient than pure transformers for small models. As the model size increases (in terms of detection inference FLOPs), pure ViTs scale better than hybrids both in overall and zero-shot performance. Pure ResNets perform poorly in our setup. Colored markers indicate the best model of a given size across all explored hyperparameters; light gray markers indicate the suboptimal hyperparameters. Asterisks ($\ast$) indicate models trained with random negative labels.
    \emph{Right:} Architecture also influences which aspects of the task a model learns: Pure ViTs perform systematically better at zero-shot detection (\lvisAPr{}) than hybrid architectures at a given overall object-level performance (\lvisAP{}). We speculate that ViTs are biased towards learning semantic generalization, whereas ResNets/Hybrids are biased towards learning localization of known classes. This difference diminishes as model size and performance increases.
    }
    \label{fig:performance_by_architecture}
    \vspace{-5mm}
\end{figure*}

Which factors contribute to strong transfer? Prior work on classification found that pre-training and model size must be scaled \emph{together} to achieve optimal transfer -- over-training small models on large data can even lead to reduced performance~\cite{kolesnikov2020big}. We find this effect to be even stronger for transfer to detection. As the amount of pre-training is increased, detection performance increases at first but then peaks, while image-level performance continues to increase (\Cref{fig:image_vs_object_level_performance}, right). However, the positive trend of detection performance with pre-training can be extended by increasing model size and improving detection fine-tuning (\Cref{fig:image_vs_object_level_performance}, right, R50+H/32).
\enlargethispage{\baselineskip}

Given that increasing model size improves performance, an important question is which architectures have the most favorable scaling properties. For classification, Transformer-based architectures have been found to be more efficient in terms of pre-training compute than ResNets, and hybrid ResNet-Transformer architectures to be the most efficient, at least at smaller computational budgets~\cite{dosovitskiy2021vit}. In addition, ResNets were found to be better when little pre-training data is available, but were overtaken by Transformers as available data increases~\cite{dosovitskiy2021vit,steiner2021train}. We performed a similar analysis for detection. 
Using detection inference compute as the measure of model size, and choosing the best hyperparameters and pre-training duration for each size, we found that hybrid models tend to be more efficient than pure ViTs at small model sizes, while ResNets perform poorly in our setup (\Cref{fig:performance_by_architecture}). However, for large models, pure ViTs overtake hybrids. 
To start explaining this difference, we compared overall and zero-shot detection performance and found a clear dissociation between hybrids and pure Transformers (at least at small model sizes; \Cref{fig:performance_by_architecture}, right). This perhaps indicates that Transformers are more biased than hybrid architectures towards learning semantic generalization (necessary for high zero-shot performance), which might be beneficial when large-scale pre-training is possible.
Overall, our findings go beyond those for classification and suggest that further scaling efforts should focus on pure Transformer architectures.

%%%%%%%%%%%%%%%%%%%%%%%

\subsection{How to Unlock Pre-Training Potential for Detection}
\label{sec:experiments:unlock-downstream-potential}

In \Cref{sec:scaling}, we found that strong image-level performance is necessary, but not sufficient, for strong detection performance. We will now describe our recipe for obtaining strong open-vocabulary detection performance after image-level pre-training. Ultimately, all components of our recipe aim at reducing overfitting on the relatively small number of available detection annotations, and the small semantic label space covered by the annotations.
Our approach relies on (i) measures to stabilize optimization, (ii) careful use of the available detection training data, and (iii) a range of data augmentations. We discuss these ablations in detail below, where numbers in italic (e.g. \textit{(15)}) refer to individual ablation experiments in Table~\ref{tab:ablations}. 
Importantly, the optimal recipe for zero-shot performance (\lvisAPr{}) does not necessarily maximize in-distribution performance (\oiAP{}). We discuss this finding and further ablations in Appendix~\ref{appendix:sec:detailed-ablations}.

\begingroup
\begin{table}[t]
    \centering
    \caption{Ablation study of the main methodological improvements necessary for successful transfer of image-text models to detection. For simplicity, difference in AP to the \textit{baseline} is shown. Except for the experiment retraining LVIS rare labels (last row), all differences are expected to be negative. To reduce variance, all results are averaged across two replicates. All ablations were carried out for the ViT-R26+B/32 model, and unless otherwise specified used a $70$K step training schedule.}
    \label{tab:ablations}
    \resizebox{0.85\textwidth}{!}{%
    \begin{tabular}{lrrrr}
    \toprule
          \textbf{Ablation} & \textbf{\lvisAP{}} & \textbf{\lvisAPr} & \textbf{\cocoAP{}} & \textbf{\oiAP{}} \\
    \midrule
    \vspace{0.5ex}\textit{Baseline}                    &        $21.0$ &        $18.9$ &         $30.9$ &  $54.1$ \\

    ~\textit{(1)}~Only use VG for training             &       $-14.5$ &        $-14.0$ &       $-23.6$ & $-38.3$ \\
    ~\textit{(2)}~Only use OI for training             &        $-6.9$ &         $-5.7$ &        $-4.2$ &   $0.3$\vspace{0.3ex}\\
    
    ~\textit{(3)}~Same LR for image and text encoders  &        $-3.0$ &         $-8.5$ &        $-0.5$ &   $0.4$\vspace{0.3ex}\\
    
    ~\textit{(4)}~No prompt ensembling at inference    &        $-2.8$ &         $-5.5$ &        $-5.9$ &  $-0.1$ \\
    ~\textit{(5)}~No prompts (train or inference)      &        $-1.2$ &         $-1.3$ &        $-0.6$ &  $-6.3$\vspace{0.3ex}\\
    
    ~\textit{(6)}~No random negatives                  &        $-1.0$ &         $-2.8$ &        $-0.4$ &   $1.0$\vspace{0.3ex}\\
    
    ~\textit{(7)}~No mosaics                           &        $-2.3$ &         $-1.5$ &        $-1.7$ &  $-0.7$ \\
    ~\textit{(8)}~No mosaics, train 2x longer          &        $-2.9$ &         $-2.8$ &        $-1.8$ &  $-0.7$ \\
    ~\textit{(9)}~No mosaics, train 3x longer          &        $-3.4$ &         $-3.6$ &        $-1.8$ &  $-0.8$\vspace{0.3ex}\\
    
    ~\textit{(10)}~Do not merge overlapping instances  &        $-0.8$ &         $-1.3$ &        $-0.6$ &  $-0.7$\vspace{0.3ex}\\
    
    ~\textit{(11)}~No location bias in box predictor   &        $-1.2$ &         $-1.1$ &        $-1.3$ &  $-1.0$\vspace{0.3ex}\\
    
    ~\textit{(12)}~Do not filter out \textit{any} cropped boxes &        $-0.1$ &          $0.0$ &         $0.1$ &  $-0.1$ \\
    ~\textit{(13)}~Filter out \textit{all} cropped boxes &        $-0.1$ &         $-0.6$ &         $0.1$ &  $0.2$\vspace{0.3ex}\\

    ~\textit{(14)}~Do not remove OI crowd instances    &         $0.0$ &          $0.7$ &        $-0.4$ &   $3.0$\vspace{0.3ex}\\
    
    ~\textit{(15)}~Do not remove LVIS rare labels      &         $0.1$ &          $0.2$ &         $-0.1$ &   $1.1$ \\
    \bottomrule
    \end{tabular}}
\end{table}
\endgroup

\subsubshift{}
\subsubsection{Stabilizing Optimization.}
The goal of fine-tuning is to learn from the available detection data without destroying the representations learned during pre-training. To this end, we take the following measures. First, we \textbf{reduce the learning rate of the text encoder} to $2\times10^{-6}$ (i.e. $100\times$ smaller than the image encoder learning rate) during fine-tuning~\textit{(3)}. This reduces overfitting, possibly by preventing the text encoder from ``forgetting'' the semantics learned during pre-training while fine-tuning on the small space of detection labels. Interestingly, freezing the text encoder completely yields poor results. Second, we \textbf{bias predicted box coordinates}~\textit{(11)} to be centred at the position of the corresponding token on the 2D grid, as described in \Cref{sec:method:architecture}. This speeds up learning and improves final performance, presumably by breaking symmetry during the bipartite matching used in the loss. Third, for larger models, we use \textbf{stochastic depth regularisation} \cite{huang2016deep,arnab2021vivit} with probability of $0.1$ on both the image and text encoders, and \textbf{shorter training schedules} (Section~\ref{appendix:sec:hyperparamters}). 
\subsubshift{}
\subsubsection{Careful Use of Available Detection Data.} As our ablations show (Table~\ref{tab:ablations}), the amount of detection training data is a limiting factor for the performance of our models. Therefore, we \textbf{combine multiple datasets} -- OI+VG for most models in our study~\textit{(1-2)}, and O365+VG for the largest models as indicated in \Cref{tab:main-results}. Further, we take care to keep the available annotations free of noise: We \textbf{remove ``group'' annotations and ``not exhaustively annotated'' categories}~\textit{(14)} from datasets indicating such annotations (e.g. OI). These annotations provide conflicting supervision to the model because it cannot learn (except through memorization) which annotations are exhaustive and which are not. Removing them improves performance of larger models. In addition, we \textbf{remove partial boxes left by random crop augmentation}, since these can also provide conflicting supervision if most of an object was actually cropped out. Retaining instances with at least $60\%$ of their original area leads to better results than retaining all \textit{(12)} or only uncropped \textit{(13)} instances.

\subsubshift{}
\subsubsection{Augmentations.} Finally, we enrich the available detection labels through augmentation of both images and queries. On the images, we use \textbf{random cropping} (removing partially cropped boxes as described above). Additionally, we use \textbf{image scale augmentation} similar to ``large scale jitter'' \cite{ghiasi2021simple}. 
However, instead of simply resizing and padding images, we tile several downscaled images into one large ``mosaic'' image. We randomly sample single images, $2 \times 2$ grids, and $3 \times 3$ grids with probabilities $0.5$, $0.33$, and $0.17$, respectively \textit{(7-9)}. To augment the queries (category names), we use \textbf{random prompts} during training, and \textbf{ensemble predictions over several prompts} for evaluation \textit{(4-5)}. We use the 80 CLIP prompts for training and ensemble over the 7 ``best'' CLIP prompts (as defined in \cite{radford2021clip}) during evaluation. 
Finally, we randomly sample \textbf{pseudo-negative labels} for each image until there are at least $50$ negative labels~\cite{zhou2021probabilistic}. Further implementation details are provided in \Cref{appendix:sec:random-negatives,appendix:sec:mosaics}.

\newcommand{\resultswidth}{1.06in}
\newcommand{\resultsspace}{\,}
\newcommand{\sidelabel}[1]{\rot{\footnotesize #1}}

\section{Conclusion}
We presented a simple recipe for transferring contrastively trained image-text models to detection. 
Our method achieves zero-shot detection results competitive with much more complex approaches on the challenging LVIS benchmark and outperforms existing methods on image-conditioned detection by a large margin. 
Our results suggest that pre-training on billions of image-text examples confers strong generalization ability that can be transferred to detection even if only relatively limited object-level data are available (millions of examples).
In our analyses we disentangle the determinants of successful transfer of image-level representations to detection, and show that pre-training simple, scalable architectures on more data leads to strong zero-shot detection performance, mirroring previous observations for image classification tasks. 
We hope that our model will serve as a strong starting point for further research on open-world detection.

\subsubsection{Acknowledgements.} We would like to thank Sunayana Rane and Rianne van den Berg for help with the DETR implementation, Lucas Beyer for the data deduplication code, and Yi Tay for useful advice.

\bibliographystyle{splncs04}
\bibliography{bibliography}

\clearpage

\appendix

\setcounter{table}{0}
\renewcommand{\thetable}{A\arabic{table}}
\setcounter{figure}{0}
\renewcommand{\thefigure}{A\arabic{figure}}

% This removes the leading A from the appendix heading:
\renewcommand{\thesection}{}
\renewcommand{\thesubsection}{A\arabic{section}.\arabic{subsection}}
\makeatletter
\def\@seccntformat#1{\csname #1ignore\expandafter\endcsname\csname the#1\endcsname\quad}
\let\sectionignore\@gobbletwo
\let\latex@numberline\numberline
\def\numberline#1{\if\relax#1\relax\else\latex@numberline{#1}\fi}
\makeatother

\section{Appendix}
The appendix provides additional examples, results and methodological details. For remaining questions, please refer to the code at \githuburl{}.

\subsection{Qualitative Examples}
\vspace{-6mm}
\begin{figure}[h]
    \centering
    \includegraphics[width=0.8\textwidth]{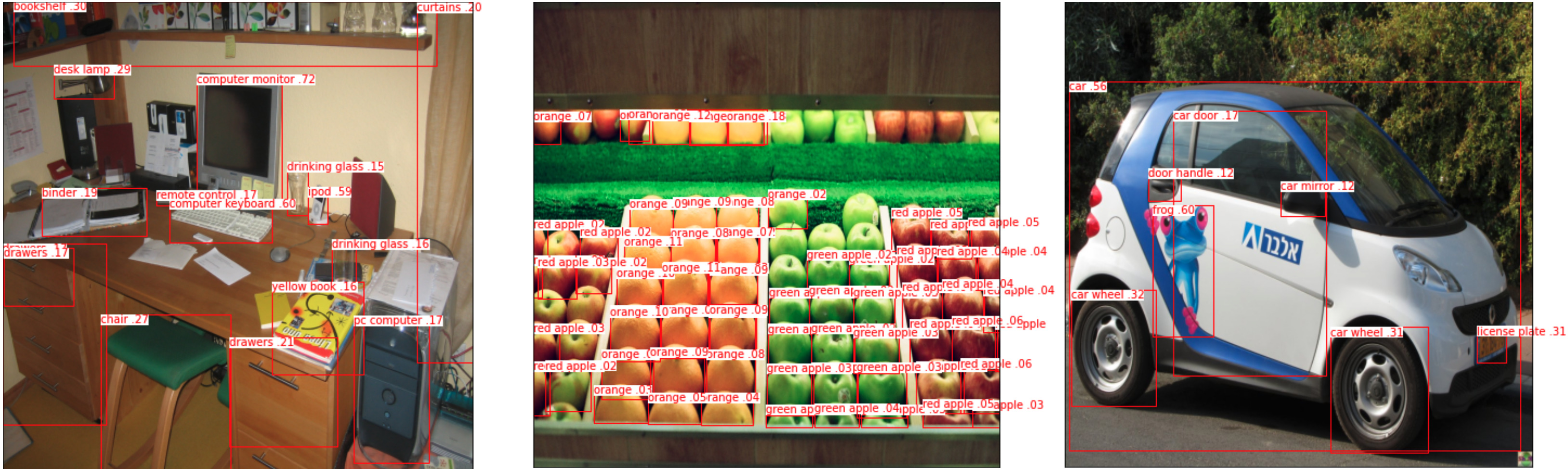}
    \caption{Text conditioning examples. Prompts: \texttt{"an image of a \{\}"}, where \texttt{\{\}} is replaced with one of 
    \texttt{bookshelf, desk lamp, computer keyboard, binder, pc computer, computer mouse, computer monitor, chair, drawers, drinking glass, ipod, pink book, yellow book, curtains, red apple, banana, green apple, orange, grapefruit, potato, for sale sign, car wheel, car door, car mirror, gas tank, frog, head lights, license plate, door handle, tail lights}.}
    \label{fig:image-conditioning-3}
\end{figure}
\enlargethispage{20mm}

\vspace{-12mm}
\begin{figure}[h!]
    \centering
    \includegraphics[width=0.8\textwidth]{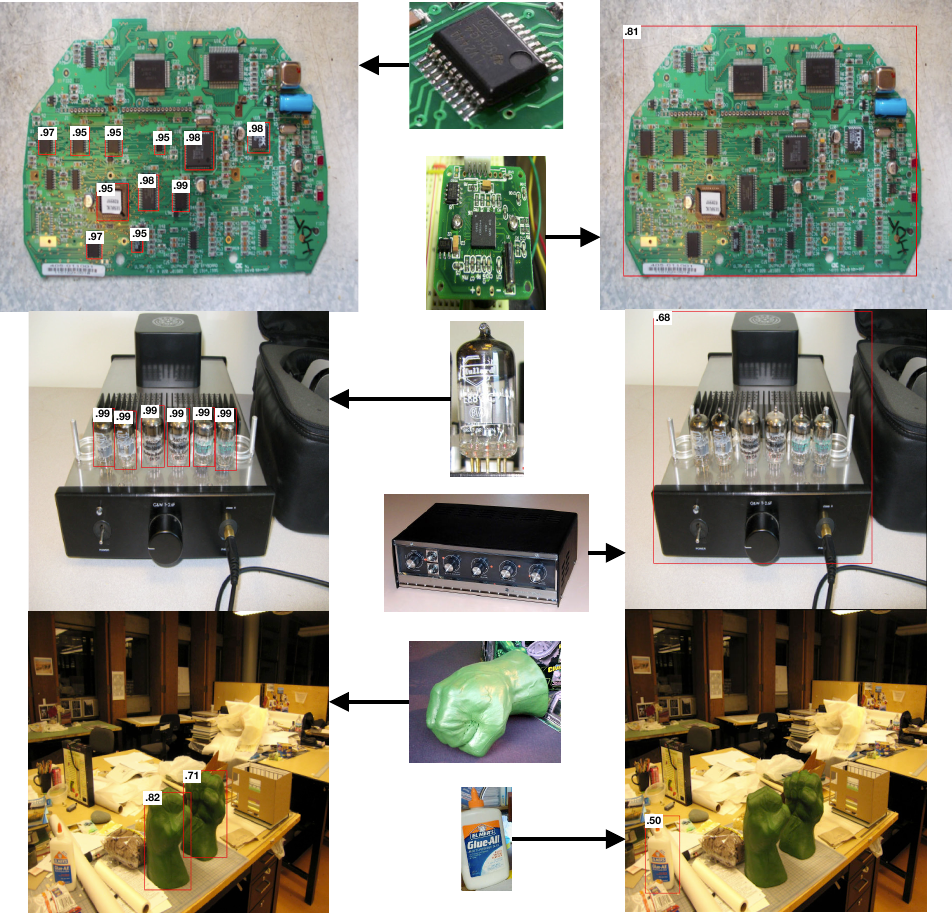}
    \caption{Image conditioning examples. The center column shows the query patches and the outer columns show the detections along with the similarity score.}    \label{fig:image-conditioning-2}
\end{figure}
\vspace{-12mm}
\clearpage

\subsection{Detection Datasets}
\label{appendix:sec:datasets}
Five datasets with object detection annotations were used for fine-tuning and evaluation in this work. Table \ref{appendix:tab:datasets} shows relevant statistics for each of these datasets:

% Sorted by year of publication.
\textbf{MS-COCO} (COCO) \cite{lin2014coco}: The Microsoft Common Objects in Context dataset is a medium-scale object detection dataset. It has about $900$k bounding box annotations for $80$ object categories, with about $7.3$~annotations per image. It is one of the most used object detection datasets, and its images are often used within other datasets (including VG and LVIS). This work uses the 2017 train, validation and test splits.

\textbf{Visual Genome} (VG)~\cite{krishnavisualgenome} contains dense annotations for objects, regions, object attributes, and their relationships within each image. VG is based on COCO images, which are re-annotated with free-text annotations for an average of $35$ objects per image. All entities are canonicalized to WordNet synsets. We only use object annotations from this dataset, and do not train models using the attribute, relationship or region annotations.

\textbf{Objects~365} (O365) \cite{shao2019vg} is a large-scale object detection dataset with $365$ object categories. The version we use has over $10$M bounding boxes with about $15.8$ object annotations per image.

\textbf{LVIS} \cite{gupta2019lvis}: The Large Vocabulary Instance Segmentation dataset has over a thousand object categories, following a long-tail distribution with some categories having only a few examples. Similarly to VG, LVIS uses the same images as in COCO, re-annotated with a larger number of object categories. In contrast to COCO and O365, LVIS is a federated dataset, which means that only a subset of categories is annotated in each image. Annotations therefore include positive and negative object labels for objects that are present and categories that are not present, respectively. In addition, LVIS categories are not pairwise disjoint, such that the same object can belong to several categories.

\textbf{OpenImages~V4} (OI)~\cite{kuznetsova2020openimages} is currently the largest public object detection dataset with about $14.6$ bounding box annotations (about $8$ annotations per image). Like LVIS, it is a federated dataset.

\begin{table}[h]
    \renewcommand{\bf}[1]{\textbf{#1}}
    \centering
    \caption{Statistics of object detection datasets used in this work.}
    \label{appendix:tab:datasets}
    \begin{tabular}{p{3cm} >{\raggedleft}p{1.5cm} >{\raggedleft}p{1.5cm} >{\raggedleft}p{1.5cm}  p{2cm}<{\raggedleft}}
    \toprule
        \bf{Name} & \bf{Train} & \bf{Val} & \bf{Test} & \bf{Categories} \\
    \midrule
        MS-COCO~2017 \cite{lin2014coco}               & $118$k    & $5$k     & $40.1$k & $80$   \\
        Visual Genome \cite{krishnavisualgenome}      & $84.5$k   & $21.6$k  & -       & -  \\
        Objects~365 \cite{shao2019vg}                 & $608.5$k  & $30$k    & -       & $365$\vspace{0.3ex}\\
        LVIS \cite{gupta2019lvis}                     & $100$k    & $19.8$k  & $19.8$k & $1203$   \\
        OpenImages~V4 \cite{kuznetsova2020openimages} & $1.7$M    & $41.6$k  & $125$k  & $601$  \\
    \bottomrule
    \end{tabular}
\end{table}

\subsubsection{De-duplication}
\label{appendix:sec:datasets-dedup}
Our detection models are typically fine-tuned on a combination of OpenImages~V4 (OI) and Visual~Genome (VG) datasets and evaluated on MS-COCO~2017 (COCO) and LVIS. In several experiments our models are additionally trained on Objects~365 (O365). We never train on COCO and LVIS datasets, but the public versions of our training datasets contain some of the same images as the COCO and LVIS validation sets. To ensure that our models see no validation images during training, we filter out images from OI, VG and O365 train splits that also appear in LVIS and COCO validation and tests splits following a procedure identical to \cite{kolesnikov2020big}. De-duplication statistics are given in Table~\ref{appendix:tab:dataset-deduplication}.

\begin{table}[h]
    \renewcommand{\bf}[1]{\textbf{#1}}
    \centering
    \caption{Train dataset de-duplication statistics. `Examples' refers to images and `instances' refers to bounding boxes.}
    \label{appendix:tab:dataset-deduplication}
    \begin{tabular}{lrrrrrr}
    \toprule
             & \multicolumn{2}{c}{\bf{Original}} & \multicolumn{2}{c}{\bf{Duplicates}} & \multicolumn{2}{c}{\bf{Remaining}} \\
     \cmidrule(l){2-3} \cmidrule(l){4-5} \cmidrule(l){6-7}
       \bf{Name} & Examples & Instances & Examples & Instances & Examples & Instances  \\
    \midrule
        OpenImages~V4  & $1.7$M & $14.6$M & $948$ & $6.4$k & $1.7$M & $14.6$M \\
        Visual~Genome  & $86.5$k & $2$M & $6.7$k & $156$k & $79.8$K & $1.9$M \\
        Objects~365    & $608.6$k & $9.2$M & $147$ & $2.4$k & $608.5$k & $9.2$M \\
    \bottomrule
    \end{tabular}
\end{table}

\subsection{Hyper-parameters}
\label{appendix:sec:hyperparamters}
Table~\ref{appendix:tab:hparams} provides an exhaustive overview of the hyper-parameter settings used for our main experiments. Beyond this, we
\begin{itemize}
    \item used cosine learning rate decay;
    \item used focal loss with $\alpha=0.3$ and $\gamma=2.0$;
    \item set equal weights for the bounding box, gIoU and classification losses \cite{carion2020detr};
    \item used the Adam optimizer with $\beta_1=0.9$, $\beta_2=0.999$;
    \item used per-example global norm gradient clipping (see Section~\ref{appendix:sec:detailed-ablations});
    \item limited the text encoder input length to $16$ tokens for both LIT and CLIP-based models.
\end{itemize}

\begingroup
\newcommand*\rot{\rotatebox{90}}
\newcommand{\sn}[1]{\num[scientific-notation=true]{#1}}
\renewcommand{\bf}[1]{\textbf{#1}}
\setlength{\tabcolsep}{1mm}

\begin{table}[t!]
   \caption{List of hyperparameters used for all models shown in the paper. Asterisks ($\ast$) indicate parameters varied in sweeps. MAP and GAP indicate the use of multihead attention pooling and global average pooling for image-level representation aggregation. Where two numbers are given for the droplayer rate, the first is for the image encoder and the second for the text encoder.}
   \label{appendix:tab:hparams}
   \centerline{
   \resizebox{1.3\textwidth}{!}{%
   \begin{tabular}{lcccccccccccccccc}
   & \rot{\bf{Training duration}} & \rot{\bf{Batch size}} & \rot{\bf{Learning rate}} & \rot{\bf{Weight decay}} & \rot{\bf{Image size}} & \rot{\bf{Pool type}} & \rot{\bf{Training steps}} & \rot{\bf{Batch size}} & \rot{\bf{Learning rate}} & \rot{\bf{Weight decay}} & \rot{\bf{Droplayer rate}} & \rot{\bf{Image size}} & \rot{\bf{Training datasets}} & \rot{\bf{Dataset proportions}} & \rot{\bf{Mosaic proportions}} & \rot{\bf{Random negatives}} \\
   \cmidrule(l){2-7} \cmidrule(l){8-17}
   \bf{Model} & \multicolumn{6}{c}{\bf{Image-level pre-training}} & \multicolumn{10}{c}{\bf{Detection fine-tuning}}\\
   \midrule
   \multicolumn{17}{l}{\emph{CLIP-based OWL-ViT models from \Cref{tab:main-results}:}}\\
      B/32 &                        &              &               &               &              &             &             140k &            256 &         \sn{5e-05} &           \sn{0} &               .2/.1 &            768 &              O365, VG &               .8/.2 &            .4/.3/.3 &                           yes \\
      B/16 &                        &              &               &               &              &             &              140k &            256 &         \sn{5e-05} &           \sn{0} &               .2/.1 &            768 &              O365, VG &               .8/.2 &             .4/.3/.3 &                           yes \\
      L/14 &                        &              &               &               &              &             &              70k &            256 &         \sn{2e-05} &           \sn{0} &               .2/.1 &            840 &              O365, VG &               .8/.2 &             .4/.3/.3 &                           yes \\
   \midrule
   \multicolumn{17}{l}{\emph{LiT-based OWL-ViT models from \Cref{tab:main-results}:}}\\
      B/32 &                     16B &            16k &         \sn{0.0003} &         \sn{1e-05} &            224 &           MAP &             140k &            256 &         \sn{0.0002} &           \sn{0} &              0.0 &            768 &              O365, VG &               .8/.2 &            .4/.3/.3 &                          yes \\
      B/16 &                      8B &            16k &         \sn{0.0003} &         \sn{1e-05} &            224 &           MAP &             140k &            256 &         \sn{0.0002} &           \sn{0} &              0.0 &            768 &              O365, VG &               .8/.2 &            .4/.3/.3 &                          yes \\
    R26+B/32 &                     16B &            16k &         \sn{0.0003} &         \sn{1e-05} &            288 &           MAP &             140k &            256 &         \sn{0.0002} &           \sn{0} &              0.0 &            768 &              O365, VG &               .8/.2 &            .4/.3/.3 &                          yes \\
      L/16 &                     16B &            16k &         \sn{0.0003} &         \sn{1e-05} &            224 &           MAP &              70k &            256 &         \sn{5e-05} &           \sn{0} &              0.0 &            768 &              O365, VG &               .8/.2 &            .4/.3/.3 &                          yes \\
      H/14 &                     12B &            16k &         \sn{0.0003} &         \sn{1e-05} &            224 &           MAP &              70k &            256 &         \sn{5e-05} &           \sn{0} &            .1/.0 &            840 &          O365, VG &              .8/.2 &             .4/.3/.3 &                           yes \\
   \midrule
   \multicolumn{17}{l}{\emph{Model used for one-shot detection (\Cref{tab:coco_one_shot}):}}\\
    R50+H/32 &                     24B &            12k &         \sn{0.0007} &         \sn{1e-05} &            224 &           GAP &              28k &            256 &         \sn{0.0002} &           \sn{0} &              0.1 &            960 &          OI, O365, VG &              .4/.4/.2 &            .5/.33/.17 &                          yes \\
   \midrule
   \multicolumn{17}{l}{\emph{Baseline models for the ablation study (\Cref{tab:ablations,appendix:tab:detailed-ablations}):}}\\
      B/32 &                      2B &            16k &         \sn{0.0003} &         \sn{1e-05} &            224 &           MAP &              70k &            256 &         \sn{0.0002} &           \sn{0} &              0.0 &            768 &              OI, VG &               .7/.3 &            .5/.33/.17 &                          yes \\
    R26+B/32 &                      8B &            16k &         \sn{0.0003} &         \sn{1e-05} &            288 &           MAP &              70k &            256 &         \sn{0.0002} &           \sn{0} &              0.0 &            768 &              OI, VG &               .7/.3 &            .5/.33/.17 &                          yes \\
   \midrule
   \multicolumn{17}{l}{\emph{Models used in the scaling study (\Cref{fig:image_vs_object_level_performance,fig:performance_by_architecture}):}}\\
      $\ast$ &                     $\ast$ &            16k &         $\ast$ &         $\ast$ &            $\ast$ &           MAP &             140k &            256 &         $\ast$ &           \sn{0} &               0.0 &            768 &              OI, VG &               .7/.3 &            .5/.33/.17 &                           no \\
    R50+H/32 &                     $\ast$ &            12k &         \sn{0.0007} &         \sn{1e-05} &            224 &           GAP &              28k &            256 &         \sn{0.0002} &           \sn{0} &              0.0 &            960 &              OI, VG &               .7/.3 &            .5/.33/.17 &                          yes \\
   \bottomrule
   \end{tabular}
   }}
\end{table}
\endgroup

\subsubsection{CLIP-based models.}
The visual encoder of the publicly available CLIP models provides, in addition to the image embedding features, a class token. In order to evaluate whether the information in the class token is useful for detection fine-tuning, we explored to either drop this token, or to merge it into other feature map tokens by multiplying it with them. We found that multiplying the class token with the feature map tokens, followed by layer norm, worked best for the majority of architectures, so we use this approach throughout. Other hyper-parameters used in the fine-tuning of CLIP models are shown in Table~\ref{appendix:tab:hparams}.

\subsection{Pre-Training Image Resolution}
\begin{figure*}[ht!]
    \centerline{
        \includegraphics[width=0.8\textwidth]{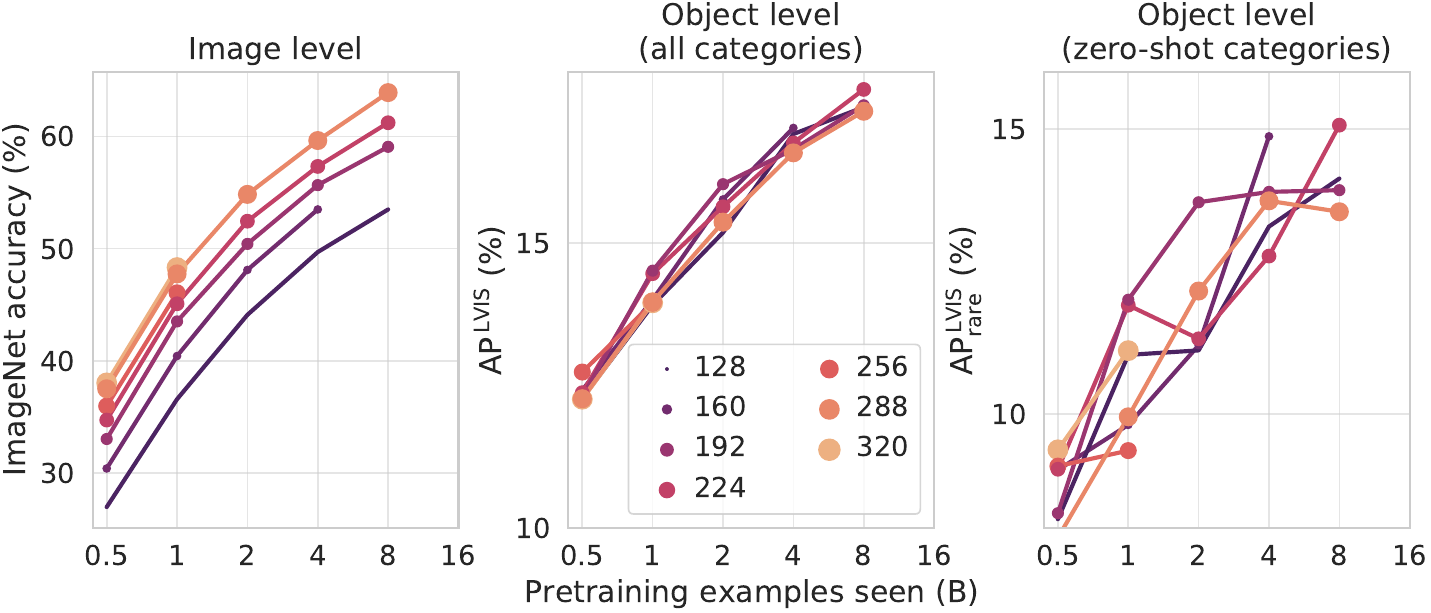}
    }
    \caption{Effect of image size used during image-level pre-training on zero-shot classification and detection performance shown for the ViT-B/32 architecture.}
    \label{appendix:fig:upstream_resolution}
\end{figure*}

We investigated the effect of the image size used during image-text pre-training, on zero-shot classification and detection performance (\Cref{appendix:fig:upstream_resolution}). To reduce clutter the results are shown for the ViT-B/32 architecture only, but the observed trends extend to other architectures, including Hybrid Transformers. The use of larger images during pre-training consistently benefits zero-shot classification, but makes no significant difference for the detection performance. We thus default to the commonly used $224\times224$ resolution for pre-training. We used $288\times288$ for some of our experiments with Hybrid Transformer models.

\subsection{Random Negatives}
\label{appendix:sec:random-negatives}

Our models are trained on federated datasets. In such datasets, not all categories are exhaustively annotated in every image. Instead, each image comes with a number of labeled bounding boxes (making up the set of \emph{positive} categories), and a list of categories that are known to be absent from the image (i.e., \emph{negative} categories). For all other categories, their presence in the image  unknown. Since the number of negative labels can be small, prior work has found it beneficial to randomly sample ``pseudo-negative'' labels for each image and add them to the annotations \cite{zhou2021probabilistic}. We follow the same approach and add randomly sampled pseudo-negatives to the real negatives of each image until there are at least 50 negative categories. In contrast to \cite{zhou2021probabilistic}, we sample categories in proportion to their frequency in the full dataset (i.e. a weighted combination of OI, VG, and potentially O365). We exclude categories from the sample that are among the positives for the given image.

\subsection{Image Scale Augmentation}
\label{appendix:sec:mosaics}
To improve invariance of detection models to object size, prior work found it beneficial to use strong random jittering of the image scale during training~\cite{ghiasi2021simple}. We use a similar approach, but follow a two-stage strategy that minimizes image padding.

First, we randomly crop each training image. The sampling procedure is constrained to produce crops with an aspect ratio between 0.75 and 1.33, and an area between 33\% and 100\% of the original image. Bounding box annotations are retained if at least 60\% of the box area is within the post-crop image area. After cropping, images are padded to a square aspect ratio by appending gray pixels at the bottom or right edge.

\begin{figure*}[t]
    \centerline{
        \includegraphics[width=0.28\textwidth]{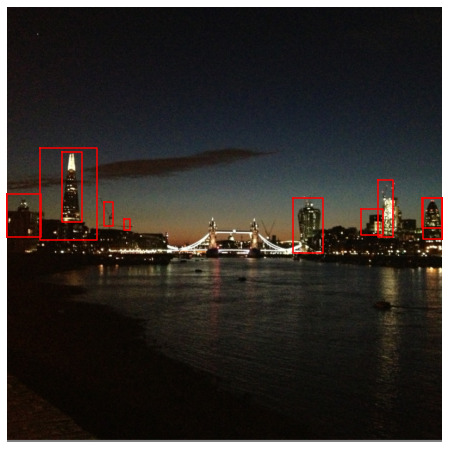}
        \hspace{3mm}
        \includegraphics[width=0.28\textwidth]{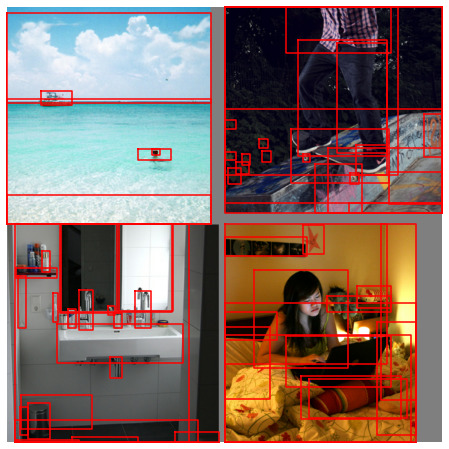}
        \hspace{3mm}
        \includegraphics[width=0.28\textwidth]{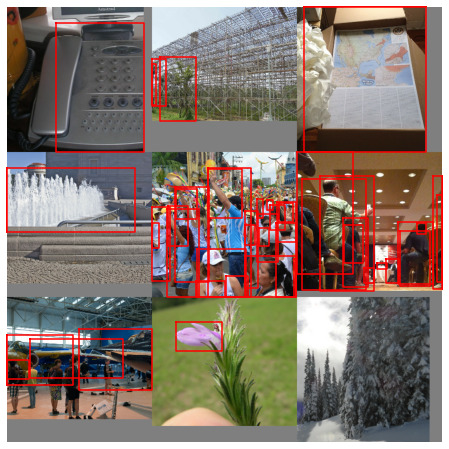}
    }
    \caption{Example training images. Ground-truth boxes are indicated in red. From left to right, a single image, a $2 \times 2$ mosaic, and a $3 \times 3$ mosaic are shown. Non-square images are padded at the bottom and right (gray color).}
    \label{fig:mosaic_examples}
\end{figure*}

Second, we assemble multiple images into grids (``mosaics'') of varying sizes, to further increase the range of image scales seen by the model. We randomly sample single images, $2 \times 2$ mosaics, and a $3 \times 3$ mosaics, with probabilities 0.5, 0.33, and 0.17, respectively, unless otherwise noted (\Cref{fig:mosaic_examples}). This procedure allows us to use widely varying images scales while avoiding excessive padding and/or the need for variable model input size during training.

\subsection{One-shot (Image-Conditioned) Detection Details}
\label{appendix:sec:image-conditioning-details}

\subsubsection{Extracting Image Embeddings to Use as Queries.}
We are given a query image patch $Q$ for which we would like to detect similar patches in a new target image, $I$. We first run inference on the image from which patch $Q$ was selected, and extract an \textit{image embedding} from our model's class head in the region of ${Q}$. In general, our model predicts many overlapping bounding boxes, some of which will have high overlap with $Q$. Each predicted bounding box $b_{i}$ has a corresponding class head feature $z_{i}$. Due to our DETR-style bipartite matching loss, our model will generally predict a single \textit{foreground} embedding for the object in $Q$ and many \textit{background} embeddings adjacent to it which should be ignored. Since all the background embeddings are similar to each other and different from the single foreground embedding, to find the foreground embedding, we search for the most \textit{dissimilar} class embedding within the group of class embeddings whose corresponding box has IoU $>0.65$ with $Q$. We score a class embedding $z_{i}$'s similarity to other class embeddings as $ f(z_{i}) = \sum_{j=0}^{N-1} z_{i} \cdot z_{j}^{T}$. Therefore, we use the most dissimilar class embedding $\mathrm{argmin}_{z_{i}} f(z_{i})$ as our query feature when running inference on $I$. In about $10$\% of the cases, there are no predicted boxes with IoU $>0.65$ with $Q$. In these cases we fall back to using the embedding for the text query \texttt{"an image of an object"}.

\subsubsection{Image-Conditioned Evaluation Protocol.}
We follow the evaluation protocol of \cite{CoAE}. During evaluation, we present the model with a target image containing at least one instance of a held-out MS-COCO category and a query image patch containing the same held-out category. Both the target image and the query patch are drawn from the validation set. We report the AP50 of the detections in the target image. Note that unlike typical object detection, it is assumed that there is at least one instance of the query image category within the target image.
Like prior work, we use Mask-RCNN~\cite{he2017maskrcnn} to filter out query patches which are too small or do not show the query object clearly. 
During detection training, we took care to hold out all categories related to any category in the held-out split. We removed annotations for any label which matched a held-out label or was a descendant of a held-out label (for example, the label ``girl'' is a descendant label of ``person''). Beyond this we also manually removed any label which was similar to a held-out category. We will publish all held-out labels with the release of our code.

\subsection{Detection results on COCO and O365}
\label{appendix:sec:coco-o365-results}
We present additional evaluation results on the COCO and O365 datasets in Table~\ref{tab:additional-results}. 
These results show the open-vocabulary generalization ability of our approach.
Although we do not train these models directly on COCO or O365 (unless otherwise noted), our training datasets contain object categories overlapping with COCO and O365, so these results are not ``zero-shot'' according to our definition.
The breadth of evaluation setups in the literature makes direct comparison to existing methods difficult. We strove to note the differences relevant for a fair comparison in Table~\ref{tab:additional-results}.

\begingroup
\setlength{\tabcolsep}{2mm}
\begin{table}[t!]
    \newcommand{\superv}[1]{\textcolor{supervisedgray}{#1}}
    \renewcommand{\check}[1]{\textcolor{red}{#1}}
    \renewcommand{\bf}[1]{\textbf{#1}}
    \centering
    \caption{Open-vocabulary detection performance on COCO and O365 datasets. 
    The results show the open-vocabulary generalization ability of our models to datasets that were not used for training.
    Results for models trained on the target dataset are shown in \superv{gray}.
    Most of our models shown here were not trained directly on COCO or O365 (they are different from the models in \Cref{tab:main-results}). However, we did not remove COCO or O365 object categories from the training data, so these numbers are not ``zero-shot''.
    For our models, we report the mean performance over three fine-tuning runs. 
    }
    \label{tab:additional-results}
    \centerline{
    \resizebox{1.3\textwidth}{!}{%
    \begin{tabular}{llccccccc}
    \toprule
    \bf{Method}                             & \bf{Backbone}         & \bf{Image-level}      & \bf{Object-level}         & \bf{Res.}         & \bf{\cocoAP{}} & \bf{\cocoAPfifty{}} & \bf{\objectsAP{}} & \bf{\objectsAPfifty{}} \\
    \midrule

    % ViLD: https://arxiv.org/pdf/2104.13921.pdf Table 5
    ViLD \cite{gu2021vild}                  & ResNet50              & CLIP                  & LVIS base                 & $1024$ & $36.6$ & $55.6$ & $11.8$ & $18.2$ \\
    % Region CLIP: https://arxiv.org/pdf/2112.09106.pdf, Table 1
    Reg.~CLIP \cite{zhong2021regionclip}     & R50-C4                & CC3M                  & COCO base                 & ? & - &  \superv{$50.4$} & - & - \\
    Reg.~CLIP \cite{zhong2021regionclip}     & R50x4-C4              & CC3M                  & COCO base                 & ? & - &  \superv{$55.7$} & - & - \\
    
    % GLIP: https://arxiv.org/pdf/2112.03857.pdf Table 3
    GLIP \cite{li2021glip}                  & Swin-T                & Cap4M                 & O365, GoldG, ...          & ? & $46.7$ & - & - & - \\
    GLIP \cite{li2021glip}                  & Swin-L                & CC12M, SBU            & OI, O365,  VG, ...        & ? & $49.8$ & - & - & - \\
    
    % Detic: https://arxiv.org/pdf/2201.02605.pdf Table 3
    Detic \cite{zhou2022detic}              & R50-C4            & CLIP, COCO-Cap            & COCO base                 & $1333$ & - & \superv{$45.0$} & - & - \\
    % Detic: https://arxiv.org/pdf/2201.02605.pdf Table 4
    Detic \cite{zhou2022detic}              & Swin-B                & CLIP, I21K            & LVIS base                 & $869$ & - & - & $21.5$ & - \\
    
    \midrule
    
    % Ours + CLIP:
    \ourmethod{} (ours)                     & ViT-B/32      & CLIP                  & OI, VG                    & $768$ & 28.1 & 44.7 & - & - \\
    \ourmethod{} (ours)                     & ViT-B/16      & CLIP                  & OI, VG                    & $768$ & $31.7$ & $49.2$ & - & - \\
    \ourmethod{} (ours)                     & ViT-L/14      & CLIP                  & O365, VG                    & 840 & 43.5 & 64.7 & - & - \\
    
    \midrule
    
    % Ours:
    \ourmethod{} (ours)                     & ViT-B/32              & LiT                   & OI, VG            & $768$ & $28.0$ & $44.4$ & $9.4$  & $15.2$  \\
    \ourmethod{} (ours)                     & ViT-B/16              & LiT                   & OI, VG            & $768$ & $30.3$ & $47.4$ & $10.7$ & $17.0$  \\
    \ourmethod{} (ours)                     & R26+B/32              & LiT                   & OI, VG            & $768$ & $30.7$ & $47.2$ & $11.1$ & $17.4$  \\
    \ourmethod{} (ours)                     & ViT-L/16              & LiT                   & OI, VG            & $672$ & $34.7$ & $53.9$ & $13.7$ & $21.6$  \\
    \ourmethod{} (ours)                     & ViT-H/14              & LiT                   & OI, VG            & $840$ & $36.0$ & $55.3$ & $15.5$ & $24.0$  \\
    \ourmethod{} (ours)                     & ViT-H/14              & LiT                   & O365, VG      & $840$ & $42.2$ & $64.5$ & - & -  \\
    \bottomrule
    \end{tabular}}}
\end{table}
\endgroup

\subsection{Extended Ablation Study}
\label{appendix:sec:detailed-ablations}

\begin{table}[t]
    \centering
    \caption{Additional ablations. VG(obj) and VG(reg) respectively refer to Visual~Genome object and region annotations.}  \label{appendix:tab:detailed-ablations}
    \resizebox{\columnwidth}{!}{
    \begin{tabular}{lrrrrrrrr}
    \toprule
          &  \multicolumn{4}{c}{\textbf{ViT-B/32}} & \multicolumn{4}{c}{\textbf{ViT-R26+B/32}} \\  
          \cmidrule(l){2-5} \cmidrule(l){6-9}
          \textbf{Ablation} &  \textbf{\lvisAP{}} & \textbf{\lvisAPr{}} & \textbf{\cocoAP{}} & \textbf{\oiAP{}} & \textbf{\lvisAP{}} & \textbf{\lvisAPr{}} & \textbf{\cocoAP{}} & \textbf{\oiAP{}} \\
    \midrule
    \vspace{0.5ex}\textit{Baseline}  &        $15.7$ &         $14.1$ &        $24.1$ &  $48.5$ &        $21.0$ &         $18.9$ &        $30.9$ &  $54.1$ \\
    
    \midrule
    \multicolumn{9}{l}{\textit{Dataset ratio.} Baseline uses OI:VG(obj) = 7:3} \\
    ~OI:VG(obj) = 2:8          &        $-1.9$ &         $-2.7$ &        $-2.4$ &  $-4.8$ &        $-4.2$ &         $-4.1$ &        $-4.7$ &  $-4.8$ \\
    ~OI:VG(obj) = 3:7          &        $-1.0$ &         $-1.9$ &        $-1.2$ &  $-3.1$ &        $-3.0$ &         $-3.0$ &        $-3.3$ &  $-2.9$ \\
    ~OI:VG(obj) = 4:6          &        $-0.6$ &         $-1.8$ &        $-0.4$ &  $-1.7$ &        $-2.2$ &         $-3.6$ &        $-2.2$ &  $-1.5$ \\
    ~OI:VG(obj) = 5:5          &         $0.0$ &         $-0.5$ &         $0.1$ &  $-0.6$ &        $-1.0$ &         $-1.1$ &        $-1.0$ &  $-1.1$ \\
    ~OI:VG(obj) = 6:4          &         $0.1$ &         $-0.6$ &         $0.1$ &  $-0.3$ &        $-0.3$ &         $-1.4$ &        $-0.4$ &  $-0.2$ \\
    ~OI:VG(obj) = 8:2          &        $-0.7$ &         $-0.9$ &        $-0.6$ &  $-0.1$ &        $-0.4$ &         $-0.3$ &         $0.2$ &   $0.4$ \\
    ~OI:VG(obj) = 9:1          &        $-1.8$ &         $-1.1$ &        $-1.6$ &   $0.1$ &        $-1.8$ &         $-1.8$ &        $-1.1$ &   $0.3$ \\
    ~OI:VG(obj, reg) = 7:3     &        $-0.6$ &          $0.0$ &        $-0.9$ &  $-3.3$ &        $-1.2$ &         $-0.5$ &        $-0.8$ &  $-3.6$ \\
    ~OI:VG(reg) = 7:3          &        $-2.1$ &         $-1.4$ &        $-2.3$ &  $-2.5$ &        $-2.9$ &         $-2.3$ &        $-2.2$ &  $-2.2$ \\
    ~Only OI                   &        $-4.9$ &         $-3.2$ &        $-3.5$ &  $-0.5$ &        $-6.9$ &         $-5.7$ &        $-4.2$ &   $0.3$ \\
    ~Only VG(obj)              &        $-8.0$ &         $-8.4$ &       $-14.2$ & $-28.5$ &       $-14.5$ &        $-14.0$ &       $-23.6$ & $-38.3$\vspace{0.5ex}\\
    
    \midrule
    \multicolumn{9}{l}{\textit{Gradient clipping.} Baseline uses per-example clipping and per-example normalization.} \\
    ~Global clip, global norm       &        $-1.0$ &         $-2.0$ &        $-1.4$ &  $-4.9$ &        $-2.3$ &         $-2.9$ &        $-2.8$ &  $-5.4$ \\
    ~Global clip, per-ex. norm      &        $-4.0$ &         $-2.6$ &        $-5.3$ &  $-4.7$ &        $-5.0$ &         $-5.0$ &        $-5.7$ &  $-5.7$\vspace{0.5ex}\\

    \midrule
    \multicolumn{9}{l}{\textit{Instance merging.} Baseline merges instance that overlap with IoU $\geq0.9$} \\
    ~No merging                &        $-0.8$ &         $-1.2$ &        $-0.3$ &  $-1.2$ &        $-0.8$ &         $-1.3$ &        $-0.6$ &  $-0.7$ \\
    ~IoU $\geq0.7$             &         $0.2$ &          $0.3$ &        $-0.2$ &   $0.1$ &         $0.2$ &          $0.2$ &         $0.0$ &   $0.6$ \\
    ~IoU $\geq0.8$             &         $0.0$ &          $0.4$ &         $0.0$ &   $0.4$ &         $0.0$ &         $-1.3$ &         $0.1$ &   $0.4$ \\
    ~IoU $\geq0.95$            &        $-0.1$ &         $-0.1$ &         $0.0$ &  $-0.7$ &        $-0.5$ &         $-1.3$ &        $-0.2$ &  $-0.5$\vspace{0.5ex}\\
    
    \midrule
    \multicolumn{9}{l}{\textit{Text encoder learning rate.} Baseline uses image LR $2\times10^{-4}$ and text LR $2\times10^{-6}$.} \\
    ~LR $2\times10^{-3}$       &        $-5.1$ &        $-10.3$ &        $-0.8$ &  $-0.6$ &        $-7.1$ &        $-14.1$ &        $-1.4$ &  $-0.5$ \\
    ~LR $2\times10^{-4}$       &        $-2.3$ &         $-6.7$ &        $-0.7$ &   $0.2$ &        $-3.0$ &         $-8.5$ &        $-0.5$ &   $0.4$ \\
    ~LR $2\times10^{-5}$       &        $-1.1$ &         $-3.8$ &        $-0.5$ &   $0.6$ &        $-1.2$ &         $-3.2$ &        $-0.4$ &   $0.9$ \\
    ~Do not fine-tune text enc. &        $-1.8$ &         $-1.2$ &        $-1.9$ &  $-0.7$ &        $-1.5$ &         $-2.3$ &        $-0.6$ &  $1.2$\vspace{0.5ex}\\

    \midrule
    \multicolumn{9}{l}{\textit{Cropped box filtering.} Baseline retains boxes with $\geq60\%$ of their original area.} \\
    ~No box area filtering     &        $-0.1$ &         $-0.3$ &        $-0.2$ &  $-0.2$ &        $-0.1$ &          $0.0$ &         $0.1$ &  $-0.1$ \\
    ~$\geq20\%$ area           &        $-0.3$ &         $-1.7$ &         $0.0$ &  $-0.3$ &        $-0.2$ &         $-0.8$ &        $-0.2$ &  $-0.1$ \\
    ~$\geq40\%$ area           &         $0.1$ &          $0.0$ &         $0.0$ &   $0.2$ &         $0.1$ &          $0.9$ &         $0.1$ &  $-0.2$ \\
    ~Only full boxes           &        $-0.2$ &         $-0.9$ &        $-0.3$ &  $-0.2$ &        $-0.1$ &         $-0.6$ &         $0.1$ &   $0.2$\vspace{0.5ex}\\

    \midrule
    \multicolumn{9}{l}{\textit{Mosaics.} Baseline uses 1-to-3-size mosaics at ratio $0.5:0.33:0.17$} \\
    ~1--2 @ 2:1                 &        $-0.4$ &         $-1.1$ &        $-0.1$ &   $0.4$ &        $-0.5$ &          $0.3$ &        $-0.5$ &   $0.0$ \\
    ~1--4 @ 4:3:2:1             &         $0.1$ &          $0.3$ &         $0.0$ &  $-0.3$ &         $0.0$ &         $-0.8$ &         $0.1$ &  $-0.3$ \\
    ~No mosaics                 &        $-1.4$ &         $-1.6$ &        $-1.5$ &  $-0.4$ &        $-2.3$ &         $-1.5$ &        $-1.7$ &  $-0.7$ \\
    ~No mosaics, 2x train sched.&        $-1.0$ &         $-1.8$ &        $-0.3$ &   $1.2$ &        $-2.9$ &         $-2.8$ &        $-1.8$ &  $-0.7$ \\
    ~No mosaics, 3x train sched.&        $-1.2$ &         $-3.4$ &         $0.3$ &   $1.1$ &        $-3.4$ &         $-3.6$ &        $-1.8$ &  $-0.8$\vspace{0.5ex}\\
    
    \midrule
    \multicolumn{9}{l}{\textit{Prompting.} Baseline uses train prompting for OI and test ensemble (ens.) prompting.} \\
    ~Train: none; test: none    &         $0.0$ &         $-0.1$ &         $0.8$ & $-10.2$ &        $-1.2$ &         $-1.3$ &        $-0.6$ &  $-6.3$ \\
    ~Train: none; test: ens.    &        $-2.6$ &         $-2.2$ &        $-7.3$ & $-11.1$ &        $-4.5$ &         $-5.0$ &       $-10.0$ &  $-6.6$ \\
    ~Train: OI+VG; test: ens.   &         $0.8$ &          $1.3$ &         $0.9$ &  $-0.1$ &        $-0.7$ &         $-0.7$ &        $-0.4$ &  $-0.2$ \\
    ~Train: VG; test: ens.      &        $-0.8$ &         $-1.1$ &        $-2.9$ &  $-7.8$ &        $-3.1$ &         $-4.0$ &        $-7.8$ &  $-5.6$\vspace{0.5ex}\\
    
    \midrule
    \multicolumn{9}{l}{\textit{Other.} Baseline uses location bias, samples $50$ random negatives and removes LVIS rare labels.} \\
    ~No location bias          &        $-2.8$ &         $-2.9$ &        $-3.7$ &  $-2.6$ &        $-1.2$ &         $-1.1$ &        $-1.3$ &  $-1.0$ \\
    ~No random negatives       &        $-1.2$ &         $-3.7$ &        $-0.8$ &  $-0.4$ &        $-1.0$ &         $-2.8$ &        $-0.4$ &   $1.0$ \\
    ~Keep LVIS rare            &         $0.1$ &          $0.9$ &         $0.0$ &   $0.7$ &         $0.1$ &          $0.2$ &        $-0.1$ &   $1.1$ \\
    \bottomrule
    \end{tabular}}
\end{table}

Table~\ref{appendix:tab:detailed-ablations} extends the ablation results provided in \Cref{tab:ablations} of the main text. It uses the same training and evaluation protocol as outlined in Table~\ref{tab:ablations}, but goes further in the range of settings and architectures (ViT-B/32 and ViT-R26+B/32) considered in the study. We discuss the additional ablations below.

\subsubsection{Dataset ratios.} In the majority of our experiments we use OI and VG datasets for training. In the ablation study presented in the main text (Table~\ref{tab:ablations}), we showed that having more training data (i.e. training on both VG and OI) improves zero-shot performance. Here, we further explored the optimal ratio in which these datasets should be mixed and found that a 7:3 = OI:VG ratio worked best. Note that this overweighs VG significantly compared to the relative size of these datasets. Overweighing VG might be beneficial because VG has a larger label space than OI, such that each VG example provides more valuable semantic supervision than each OI example. 

We also tested the relative value of VG ``object'' and ``region'' annotations. In VG, ``region'' annotations provide free-text descriptions of whole image regions, as opposed to the standard single-object annotations. Interestingly, we found that training on the region annotations hurts the generalization ability of our models, so we do not use them for training.

\subsubsection{Loss normalization and gradient clipping.} In its official implementation, DETR \cite{carion2020detr} uses \textit{local} (i.e. per-device) loss normalization and is thus sensitive to the (local) batch size. We found this to be an important detail in practice, which can significantly affect performance. We explored whether normalizing the box, gIoU and classification losses by the number of instances in the image or the number of instances in the entire batch performed better. Our experiments show that per-example normalization performs best, but only \textit{when combined with per-example gradient clipping}, i.e. when clipping the gradient norm to 1.0 for each example individually, before accumulating gradients across the batch. We found that per-example clipping improves training stability, leads to overall lower losses and allows for training models with larger batch sizes.

\subsubsection{Instance merging.} Federated datasets such as OI have non-disjoint label spaces, which means that several labels can apply to the same object, either due to (near-)synonymous labels (e.g. ``Jug'' and ``Mug''), or due to non-disjoint concepts (e.g. ``Toy'' and ``Elephant'' labels both apply to a toy elephant). Due to the annotation procedure, in which a single label is considered at a time, one object can therefore be annotated with several similar (but not identical) bounding boxes. We found it helpful to merge such instances into a single multi-label instance. Multi-label annotations are consistent with the non-disjoint nature of federated annotations and we speculate that this provides more efficient supervision to the models, since it trains each token to predict a single box for all appropriate labels. Without this instance merging, the model would be required to predict individual boxes for each label applying to an object, which clearly cannot generalize to the countless possible object labels.

To merge overlapping instances we use a randomized iterative procedure with the following steps for each image:
\renewcommand{\labelenumii}{\theenumii}
\renewcommand{\theenumii}{\theenumi.\arabic{enumii}.}
\begin{enumerate}
    \item Pick the two instances with the largest bounding box overlap.
    \item If their intersection over union (IoU) is above a given threshold:
    \begin{enumerate}
        \item  Merge their labels.
        \item  Randomly pick one of the original bounding boxes as the merged instance bounding box.
    \end{enumerate}
\end{enumerate}
The picked instances are then removed and the procedure is repeated until no instances with a high enough IoU are left.
Having explored multiple IoU thresholds, we note that not merging instances with highly similar bounding boxes is clearly worse than merging them; and that a moderately high threshold of $0.7$-$0.9$ works best in practice.

\subsubsection{Learning rates.} In Table~\ref{tab:ablations} we show that using the same learning rate for the image and text encoders is clearly sub-optimal, and that it is necessary to training the text encoder with a lower learning rate. This may help to prevent catastrophic forgetting of the wide knowledge the model acquired during the contrastive pre-training stage. Here we explore a range of text encoder learning rates and demonstrate that the learning rate for the text encoder needs to be much lower (e.g. $100\times$) than that of the image encoder to get good zero-shot transfer (\lvisAPr{}). However, freezing the text encoder completely (learning rate $0$) does not work well either. \oiAP{}, which measure in-distribution performance, behaves in the opposite way. While using the same learning rate for the image and text encoders results in a big drop in \lvisAPr{}, it increases \oiAP{}. This demonstrates that the optimal recipe for zero-shot transfer (\lvisAPr{}) does not necessarily maximize in-distribution performance (\oiAP{}).

\subsubsection{Cropped bounding box filtering.} We use random image crop augmentation when training our models. Upon manual inspection of the resulting images and bounding boxes we noticed a frequent occurrence of instances with degenerate bounding boxes that no longer matched their original instance label (e.g. a bounding box around a hand with label ``Person'' resulting from cropping most of the person out of the image). To reduce the chance of our models overfitting due to having to memorize such instances, we remove object annotations if a large fraction of their box area falls outside of the random crop area. The optimal area threshold lies between 40\% and 60\%, and that neither keeping all boxes, nor keeping only uncropped boxes, performs as well (Tables~\ref{tab:ablations} and~\ref{appendix:sec:detailed-ablations}).

\subsubsection{Mosaics.} As described in \Cref{appendix:sec:mosaics}, we perform image scale augmentation by tiling multiple small images into one large ``mosaic''. We explored mosaic sizes up to $4 \times 4$, and found that while using only $2 \times 2$ mosaics in addition to single images is clearly worse than also including larger mosaics, for the considered resolutions and patch sizes the benefits of using larger mosaics (i.e. smaller mosaic tiles) saturates with the inclusion of $3 \times 3$ or $4 \times 4$ mosaics. We have not performed extensive sweeps of the mosaic ratios, and for mosaics with grid sizes from $1 \times 1$ (i.e. a single image) to $M \times M$ we use a heuristic of sampling $k \times k$ girds with probability $\frac{2\cdot(M - k + 1)}{M\cdot(1 + M)}$, such that smaller mosaics are sampled more frequently than the larger mosaics proportionally to the mosaic size.

\subsubsection{Prompting.} 
For generating text queries, similar to prior work, we augment object category names with prompt templates such as \texttt{"a photo of a \{\}"} (where \texttt{\{\}} is replaced by the category name) to reduce the distribution shift between image-level pre-training and detection fine-tuning.
We use the prompt templates proposed by CLIP \cite{radford2021clip}.
During training, we randomly sample from the list of $80$ CLIP prompt templates such that, within an image, every instance of a category has the same prompt, but prompt templates differ between categories and across images. During testing, we evaluate the model for each of the ``7 best'' CLIP prompts and ensemble the resulting predicted probabilities by averaging them.
The results in Table~\ref{appendix:tab:detailed-ablations} show that not using any prompting does not perform well, especially on the in-distribution \oiAP{} metric. 
Perhaps unsurprisingly, test-time prompt ensembling works better in cases when random prompting was also used during training. 
In some cases, prompting can have different effects on different model architectures. For example, applying random prompt augmentation to the VG dataset tends to improve performance of the B/32 model, but worsens that of the R26+B/32 model. We speculate that this variability is due to the relatively small number of prompt templates; expanding the list of prompt templates might provide more consistent benefits. We thus only use train-time random prompting for the OI dataset, where it yields consistent benefits. 

\subsubsection{Location bias.} As discussed in the main text, biasing box predictions to the location of the corresponding image patch improves training speed and final performance. The gain is especially large for the pure Transformer architecture (ViT-B/32 in Table~\ref{appendix:sec:detailed-ablations}), where removing the bias reduces performance by almost 3 points on \lvisAP and \lvisAPr, whereas the hybrid R26+B/32 drops by only slightly more than 1 point. We therefore speculate that the spatial inductive bias of the convolutional component of the hybrid serves a similar function as the location bias.

\end{document}